\documentclass{article}

\usepackage{microtype}
\usepackage{graphicx}
\usepackage{subcaption}
\usepackage{booktabs} 

\usepackage{hyperref}

 \usepackage[preprint]{icml2026}

\usepackage{amsmath}
\usepackage{amssymb}
\usepackage{mathtools}
\usepackage{amsthm}
\usepackage{multirow}
\usepackage{algorithmic} 

\usepackage[capitalize,noabbrev]{cleveref}

\theoremstyle{plain}

\theoremstyle{definition}

\theoremstyle{remark}

\usepackage[textsize=tiny]{todonotes}

\icmltitlerunning{ShortcutBreaker: Low-Rank Noisy Bottleneck and Frequency Filtering Block for Multi-Class Unsupervised Anomaly Detection}

\begin{document}
	
	\twocolumn[
	\icmltitle{ShortcutBreaker: Low-Rank Noisy Bottleneck and Frequency Filtering Block for Multi-Class Unsupervised Anomaly Detection}
	
	
	
	\icmlsetsymbol{equal}{*}
	
	\begin{icmlauthorlist}
	\icmlauthor{Peng Tang}{TUM}
	\icmlauthor{Xiaobin Hu}{NUS}
	\icmlauthor{Yang Nan}{IC}
	\icmlauthor{Tingcheng Li}{SZUST}
	\icmlauthor{Tobias Lasser}{TUM}
	\icmlauthor{Hongwei Bran Li}{NUS}
\end{icmlauthorlist}

\icmlaffiliation{TUM}{TUM}
\icmlaffiliation{NUS}{NUS}
\icmlaffiliation{IC}{imperial college London}
\icmlaffiliation{SZUST}{Suzhou University of Science and Technology}

\icmlcorrespondingauthor{Xiaobin Hu}
	
	\icmlkeywords{Machine Learning, ICML}
	
	\vskip 0.3in
	]
	
	
	
	\printAffiliationsAndNotice{}  
	
	\begin{abstract}
        Multi-class Unsupervised Anomaly Detection (MUAD) has attracted increasing research interest for its ability to develop a unified model across multiple classes, significantly reducing the computational cost of training separate models for distinct objects. However, despite the performance gains offered by advanced Transformer-based architectures, the identity shortcut issue persists. These models tend to directly copy inputs to outputs, narrowing the reconstruction error gap between normal and abnormal cases and hindering distinguishability. To address this, we propose ShortcutBreaker, a framework designed to disrupt such shortcuts through two core innovations. First, we introduce a Low-Rank Noisy Bottleneck (LRNB) derived from matrix rank inequality, which theoretically prevents trivial identity mapping by perturbing features across two distinct reconstruction groups. Second, we design a Frequency-Filter Unit (FFU) that integrates frequency spectral filtering and global-local masking. This mechanism further enlarges the anomaly score gap, with less computational cost than standard self-attention. Extensive evaluations on four benchmarks (MVTec-AD, ViSA, Real-IAD, and Universal Medical) demonstrate the superiority of ShortcutBreaker. 
        Our code will be released.
\end{abstract}
	\section{introduction}

\begin{figure}  
		\centering
		\includegraphics[width=0.98 \linewidth]
        {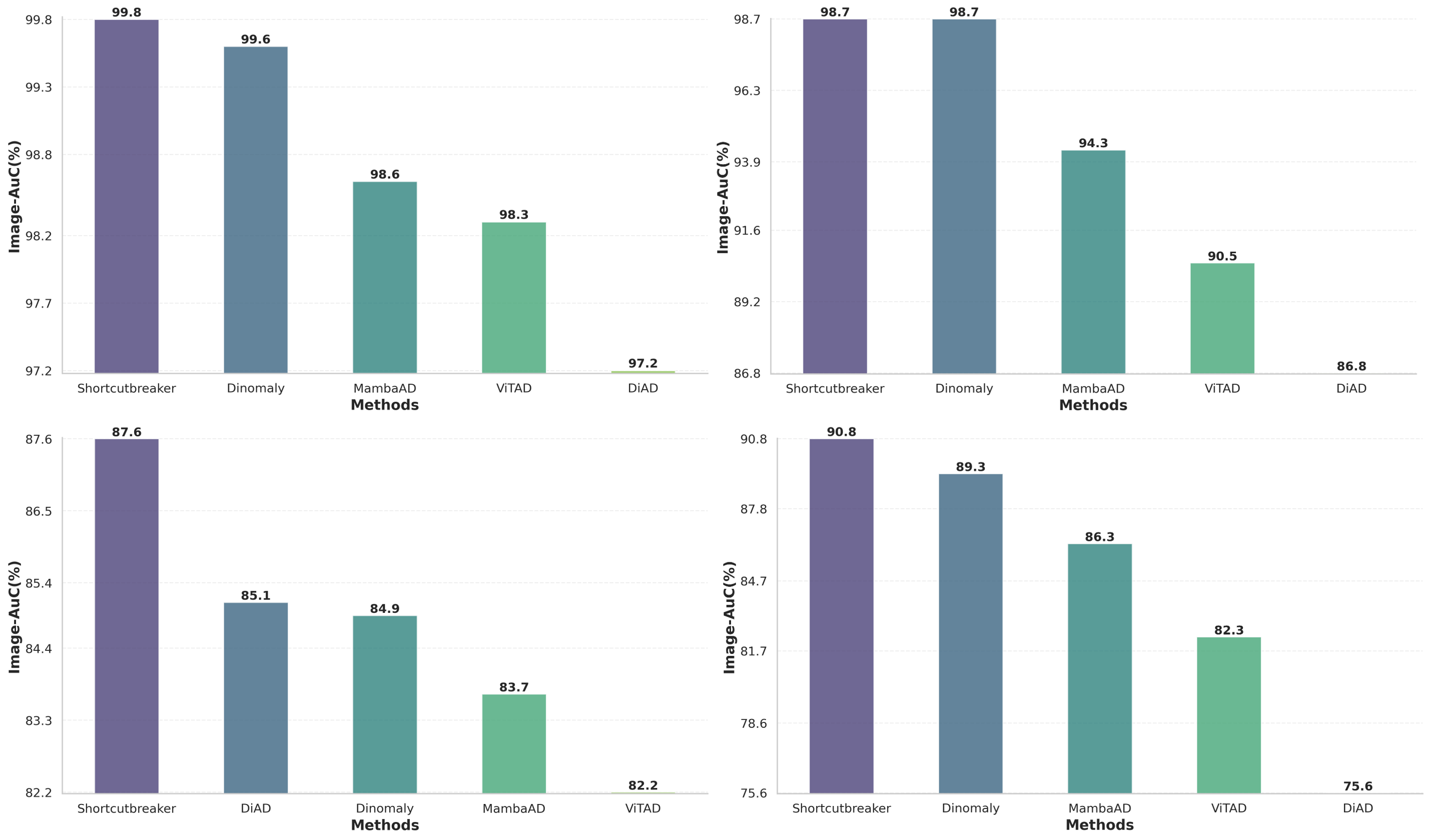}	
		
		\caption{Performance comparison in terms of image-level AUC on MVTec, ViSA, Universal Medical and Real-IAD.}
		\label{fig0_new}
		\vspace{-0.3in}
\end{figure}

	Anomaly detection in industrial and medical imaging aims to identify abnormal patterns among normal cases, saving labor and time in collecting and labeling anomalies. Given the abundance of normal cases and the scarcity of anomalies, this task is typically tackled via an unsupervised paradigm using only normal training samples.
	
	Before deep learning, anomaly detection relied on traditional techniques: density-based methods \cite{breunig2000lof,guan2015slof},distance-based methods~\cite{knorr2000distance,angiulli2005distance}, and statistics-based methods~\cite{hido2011statistical,rousseeuw2011robust}.
	Current state-of-the-art methods employ deep learning networks pre-trained on ImageNet~\cite{deng2009imagenet} to capture discriminative features. Feature reconstruction methods~\cite{guo2023encoder,guo2024recontrast,deng2022anomaly} reconstruct encoder-extracted features, assuming accurate reconstruction of normal regions but failure on unseen anomalies. 
    Memory matching methods~\cite{yi2020patch,defard2021padim,roth2022towards} memorize training-set normal features for inference matching, with those in \cite{defard2021padim,roth2022towards} using pre-trained encoders for discriminative features. 
    Pseudo-anomaly methods~\cite{liu2023simplenet,li2021cutpaste,zavrtanik2021draem} convert UAD to a supervised task by generating pseudo anomalies via noise addition to normal images/features. Hybrid methods~\cite{tien2023revisiting,zhao2023ae} integrate normalizing flows \cite{zhao2023ae} or pseudo noise \cite{tien2023revisiting} into feature reconstruction for UAD.
	
	Despite their success, these methods are limited to a one-model-one-class setup, requiring substantial storage for per-class models—especially with many disease types~\cite{you2022unified}. To address this, UniAD and follow-ups propose unified models for multi-class unsupervised anomaly detection (MUAD). However, identity mapping often emerges here, where the model returns input copies regardless of normality, enabling effective reconstruction of even anomalous samples and hindering detection~\cite{you2022unified}.
	
	\begin{figure} 
		\centering
		\includegraphics[width=1 \linewidth]{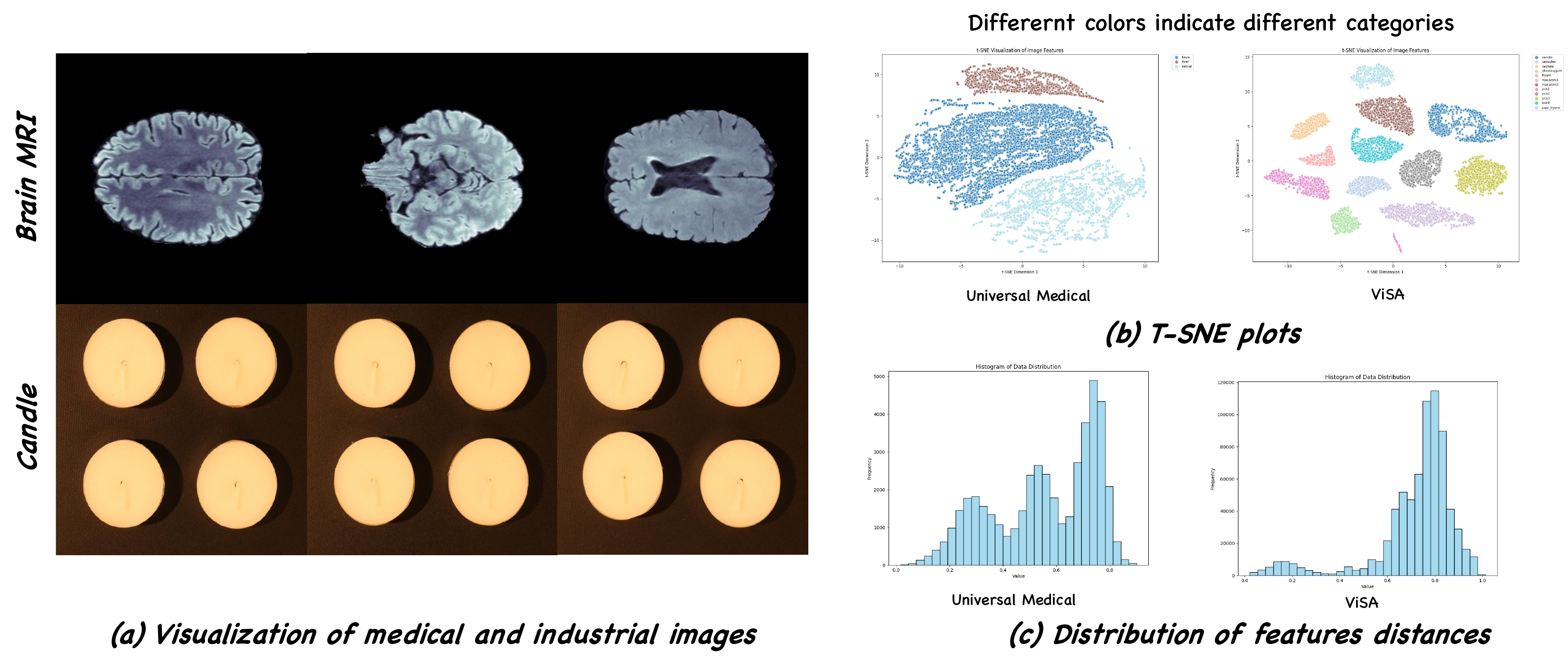}	
		\caption{The visual patterns in the medical field are richer than those in the industrial field. (a) Visualization results (b) T-SNE plots \cite{van2008visualizing}, and (c) distributions of feature distances for the Universal Medical and ViSA datasets \cite{zou2022spot} are presented. These experimental results help observe the diversity of each dataset. For T-SNE plots, we extract the final features from the pre-trained ResNet-50 \cite{he2016deep}. Moreover, we use LPIPs \cite{zhang2018unreasonable} to compare feature distances between individual image pairs.} 
		\vspace{-0.3in}
		\label{fig1_new}
	\end{figure}

	Over the past three years, substantial progress has been made in multi-class unsupervised anomaly detection (MUAD), with explorations into pretrained vision transformers (ViTs)~\cite{zhang2023exploring,guo2025dinomaly}, state space models (Mambas)~\cite{he2024mambaad}, diffusion models~\cite{yin2023lafite,he2024diffusion}, and other approaches~\cite{guo2024recontrast,zhao2023omnial,he2024learning}. Nevertheless, the multi-class setting inevitably induces identity mapping in most methods, resulting in performance degradation.
	While previous studies \cite{you2022unified,guo2025dinomaly} have endeavored to mitigate shortcut learning, the efficacy of their proposed techniques in complex and highly diverse scenarios remains limited (see Fig. \ref{fig0_new} (c)-(d)). This is particularly evident in datasets such as Real-IAD and Universal Medical, where the abundance of normal patterns exacerbates the aforementioned issue~\cite{you2022unified}. The effectiveness of simple noise operations \cite{you2022unified,guo2025dinomaly} is diminished due to the enhanced noise robustness acquired through exposure to diverse visual patterns.
    Consequently, to enhance performance, our work aims to efficiently address the problem of identity shortcuts.
	Furthermore, why are the Real-IAD and Universal Medical datasets more complex? We elaborate on this as follows: It is apparent that in comparison to widely adopted industrial datasets—MVTec-AD (encompassing 15 classes) \cite{bergmann2019mvtec} and ViSA (comprising 12 classes) \cite{zou2022spot}—Real-IAD \cite{wang2024real} exhibits more diverse normal patterns, featuring 30 object categories and 5 camera views.  
	In contrast to standardized industrial images, although the Universal Medical dataset \cite{he2024diffusion} consists of only 3 categories, the inherent heterogeneity within medical normal samples enriches the visual features (\textit{e.g.}, variations in organ size and shape across different demographics, as illustrated in Fig. \ref{fig1_new}(a)). Moreover, the Universal Medical dataset displays a sparser t-SNE plot and a wider dispersion of feature distances compared to ViSA (see Fig.~\ref{fig1_new} (b) and (c)), indicating a higher degree of intrinsic diversity within medical data. 
	
	In this paper, we propose a simple yet effective framework for the MUAD task, named ShortcutBreaker, built on the advanced DINO-pretrained vision transformer with two key innovations.
	First, drawing on two observed properties of low-rank matrix decomposition, we design a low-rank noisy bottleneck (LRNB) to effectively mitigate the identity mapping issue. Within LRNB, the low-rank property theoretically circumvents shortcut learning of unseen patterns, while learnable matrix parameters are optimized to reconstruct normal patterns.
    Second, we introduce a Frequency-Filter Unit (FFU) to replace self-attention, which efficiently curbs information leakage from input to output in the frequency domain. In FFU, a frequency spectral filtering and a local masking mechanism are integrated, forcing the decoder to learn reconstruction based on longer-range and impaired features.
	
    To validate the effectiveness of the proposed method, we conduct extensive experiments on four publicly available datasets: MVTec-AD~\cite{bergmann2019mvtec}, ViSA~\cite{zou2022spot}, Universal Medical~\cite{he2024diffusion}, and Real-IAD~\cite{wang2024real}. As presented in Fig.~\ref{fig0_new}, our ShortcutBreaker achieves the highest image-level AUC: 99.8\%, 98.9\%, 88.2\%, and 90.6\% on these four datasets, respectively. Notably, on the complex Universal Medical and Real-IAD datasets, it outperforms previous methods by a significant margin.
	In summary, our contributions are:
	\begin{itemize}
		\item We observe and simulate the properties of matrix decomposition to design a low-rank noisy bottleneck, efficiently suppressing identity mapping.
		\item We propose a frequency-filtering unit that integrates frequency spectral filtering and global-local masking operations, detecting anomalies more efficiently than standard self-attention.
		\item Our ShortcutBreaker consistently outperforms previous methods across all four datasets,demonstrating enhanced robustness in diverse scenarios.
	\end{itemize}

	\subsection{Multi Class UAD}
	UniAD~\cite{you2022unified} pioneers the field of multi-class unsupervised anomaly detection (MUAD), presenting a unified model capable of detecting anomalies across various classes.
	Most subsequent works have explored advanced modules to build better reconstruction models for MUAD. For example, LafitE~\cite{yin2023lafite} and DiAD~\cite{he2024diffusion} further advance the MUAD task by leveraging the generative power of diffusion models to better capture anomalies across multiple classes.
	ViTAD \cite{zhang2023exploring} and MambaAD \cite{he2024mambaad} develop feature reconstruction-based MUAD methods using recently advanced modules: Vision Transformer \cite{zhang2023exploring} and State Space Model \cite{he2024mambaad}, respectively.
	Few methods explicitly aim to address the identity mapping issue. UniAD~\cite{you2022unified} counteracts this through techniques like feature jittering and neighbor-masked attention. Dinomaly \cite{guo2025dinomaly} employs components such as noisy bottleneck dropout to disrupt input feature replication.
	However, these noise operations have limited effectiveness in preventing shortcut learning when training on complex medical/industrial datasets (Universal Medical and Real-IAD).

		\section{Method}
	\label{sec:method}
	\subsection{Overview}
\begin{figure*}[t]
		\centering
		\includegraphics[width=0.92 \linewidth]{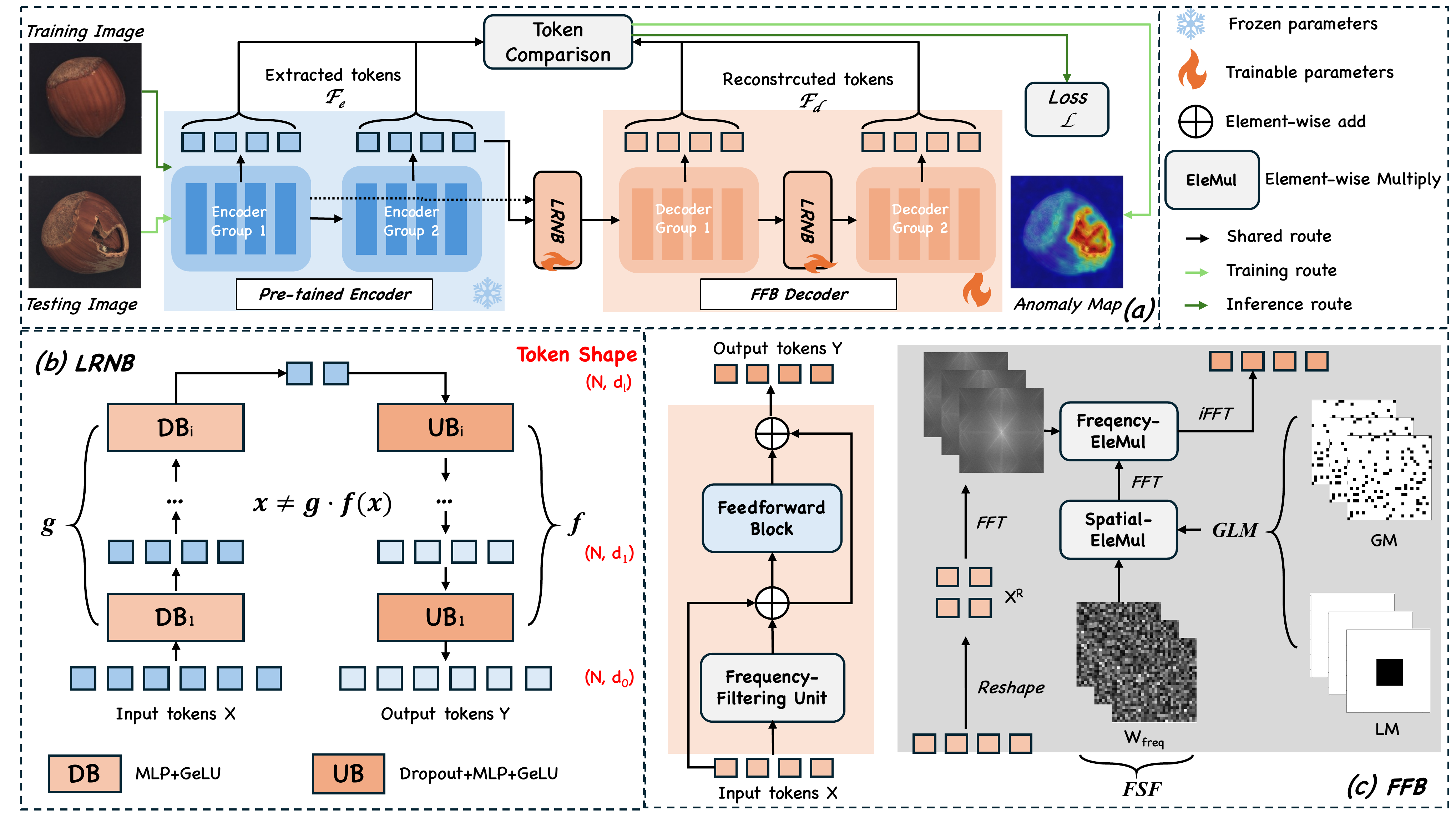}	
		\caption{Flowchart of our ShortcutBreaker: (a) the overall structure of our proposed method; (b) the structure of the Low-Rank Noisy Bottleneck (LRNB); (c) the structure of the Frequency-Filtering Block (FFB), which consists of a Frequency-Filtering Unit and a Feedforward block. FFT: Fast Fourier Transform, iFFT: Inverse Fast Fourier Transform, FSF: frequency spectral filtering, LM: local masking $M_{l}$, and GM: global masking $M_{g}$.
		}
		\vspace{-0.2in}
		\label{fig2_new}
	\end{figure*}

	Following ViTAD and Dinomaly \citep{zhang2023exploring, guo2025dinomaly}, our ShortcutBreaker is a feature-reconstruction framework and is constructed based on the vision transformer (ViT) structure.
	As depicted in Fig.~\ref{fig2_new}(a), ShortcutBreaker consists of a pre-trained encoder, a bottleneck, and a decoder.
	The DINO-pretrained ViT model \citep{darcet2023vision} is utilized as the encoder, capturing informative feature tokens to facilitate subsequent reconstruction. The bottleneck is built upon a multi-layer perceptron (MLP), integrating multi-scale feature representations from intermediate encoder layers. The decoder based on frequency-filtering blocks maintains structural similarity with the encoder, employing transformer layers to reconstruct feature maps.
	As illustrated in Fig. \ref{fig2_new}(d), the encoder remains frozen during training, while the remaining components of ShortcutBreaker are optimized to reconstruct the encoder's feature tokens by minimizing inter-token representation discrepancies. During inference, this configuration enables accurate reconstruction of normal patterns but exhibits reconstruction failures in anomalous regions. The anomaly localization map is generated by computing \(\text{1-CS}\) (where \(\text{CS}\) denotes cosine similarity) between the original and reconstructed feature tokens and then reshaping the result. The anomaly detection score is subsequently determined by the maximum value within this anomaly map.
	
	\subsection{Low-Rank Noisy Bottleneck (LRNB)}
	\label{sec: LRNB}
	As discussed in the introduction, under the MUAD setting, identity shortcuts readily occur, narrowing the gap in anomaly scores between normal and abnormal cases. Previous methods~\citep{guo2025dinomaly, you2022unified} propose perturbing extracted feature tokens to compel the network to reconstruct from information-impaired normal features, rather than directly copying them.
	However, these noisy operations are often limited in effectively preventing shortcut learning in complex scenarios.
	We hypothesize that richer visual patterns inherent in the data enhance the reconstruction model's robustness to stochastic noise. Consequently, we aim to design a novel paradigm for avoiding shortcuts that goes beyond solely relying on noisy operations.
	Through empirical observation of low-rank matrix decomposition in LoRA~\citep{hu2022lora} (or the encoder-decoder structure in Auto-Encoder~\citep{he2022masked}), we identified two inherent properties particularly suitable for addressing the identity mapping issue.
	
	
	\textbf{Property 1} This property stems from the constrained low-rank latent space, which effectively circumvents identity shortcuts. 
    In deep learning, nonlinear activations are generally used to improve performance, so we introduce the tool of the Jacobian matrix \citep{goodfellow2016deep} to conduct a linear approximation for the proof of this property. Given that the feature tokens extracted from DINO-v2 are highly informative, we assume them to be full-rank.
	
	To achieve identity mapping of $x \in \mathbb{R}^d$, we require that $ g \circ f(x) = x $ holds for all  $ x \in \mathbb{R}^n $. Differentiating both sides with respect to $x$ yields the Jacobian equation:
	\begin{equation}
		J_g(f(x)) \cdot J_f(x) = I_{d\times d}
	\end{equation}
	where $ J_f(x) \in \mathbb{R}^{d \times k} $ and $ J_g(f(x)) \in \mathbb{R}^{k \times d} $.
	Taking the determinant of both sides:
	\begin{equation}
		\label{eq8}
		r(\left(J_g(f(x)) \cdot J_f(x)\right)) = r(I_{d\times d}) = d.
	\end{equation}
	Since the product $ J_g(f(x)) \cdot J_f(x) $ is an $ d \times d $ matrix, applying the rank inequality yields:
	\begin{equation}
		\label{eq9}
		{\tiny
			r\left(J_g(f(x)) \cdot J_f(x)\right) \leq \min\left(r(J_g(f(x))), r(J_f(x))\right) \leq k}
	\end{equation}
	Here,  $r(\cdot)$ indicate the rank of the input.
	Under the low-rank adaptation module, we have  $k < d$, Substituting into Eq.~\eqref{eq9} implies:
	\begin{equation}
		\label{eq10}
		{\tiny
			r\left(J_g(f(x)) \cdot J_f(x)\right) \leq k < d}
	\end{equation}
	This contradicts Eqs.~\eqref{eq8} that $det(I_{d\times d})$ requires rank=$d$. Consequently, no such functions  $f$ and $g$ exist.

%
	
	\textbf{Property 2} Property 1 prevents the identity mapping of all the inputs; however, our final goal is a favorable score trade-off between normal and abnormal samples.
    The property 2 emerges from training exclusively on normal samples, during which the learnable \(g\) and \(f\) layers are optimized to specifically enhance the reconstruction of normal patterns. This optimization process minimizes adverse effects on the reconstruction fidelity of normal cases, thereby enlarging the gap of reconstruction errors between the two types of samples.
	

    Specifically, functions $g$ and $f$ are parameterized by Multi-Layer Perceptrons (MLPs) with GeLU activations. 
    As illustrated in Fig.~\ref{fig2_new}, $g$ and $f$ feature symmetric architectures consisting of $L$ downsampling (DB) and upsampling blocks (UB), respectively. Within these modules, each DB reduces the token dims by low-rank ratio $r$, (i.e., $d_{i} = d_{i-1} // r $, sequence length $n$ remains unchanged), while UBs correspondingly scale it back. To reinforce feature-level perturbation, Dropout~\citep{guo2025dinomaly} is integrated into the UBs. Furthermore, departing from existing methods that employ a single noise bottleneck, we deploy dual LRNBs—injecting one before each of the two decoder groups—to perturb reconstruction.
    
    \subsection{Frequency-Filtering Block}
    The self-attention mechanism is the core component of Transformer, as shown in Eq.~\ref{eq1}, enabling the model to focus on different parts of the input tokens.
    \begin{equation}\label{eq1}
    \text{Attention}(Q, K, V) = \text{Softmax}\left(\frac{QK^T}{\sqrt{d}}\right)V = A(X)V
    \end{equation} 
	where $Q\in\mathbb{R}^{N\times d}$, $K \in \mathbb{R}^{N\times d}$ and $V \in \mathbb{R}^{N\times d}$ are projected from input tokens $X \in \mathbb{R}^{N\times d}$, indicating the query, key and value vectors respectively. $K^T \in \mathbb{R}^{d\times N}$ is the transpose matrix of $K$ and $d_k$ is the scaling factor.

    In anomaly detection, Transformer blocks have recently been adopted to replace conventional convolution blocks due to their global modeling capability \cite{you2022unified, zhang2023exploring, guo2025dinomaly}. 
    However, the self-attention mechanism often learns a "spiky" weight distribution, primarily focusing on the query's own location, which facilitates an identity shortcut \cite{you2022unified, guo2025dinomaly}.

    Motivated by three intuitive benefits, we construct a Frequency-Filtering Block (FFB)  by replacing the self-attention mechanism by a Frequency-Filtering Unit, as shown in Fig.~\ref{fig2_new}(c).
    \textbf{First}, the FFU processes token sequences in the frequency domain using the 2D DFT. According to the properties of the DFT, every spectral coefficient encapsulates information from all spatial locations. By performing mixing in this spectral space, the FFU inherently maintains a holistic global modeling capability, forcing the model to reconcile inputs with a global distribution. This effectively prevents the network from relying on isolated localized shortcuts for reconstruction. 
    \textbf{Second}, unlike standard self-attention $A(X)=\text{Softmax}(QK^T/\sqrt{d})$, where input-dependent $Q$ and $K$ matrices are prone to optimizing for increased spectral norms—causing $A(X)$ to collapse into an identity matrix ($A(X)\approx I$) and creating an identity shortcut for anomalies—the FFU mitigates this issue through its input-independent architecture. The spectral filter (see $W_{freq}$ in Fig.~\ref{fig2_new}) consists of learnable parameters that remain fixed during inference. 
    During training, it learns a static spectral template representing the global energy distribution of normal patterns and cannot dynamically "re-focus" its weights on unseen anomalous signals.
    \textbf{Third}, while self-attention incurs a quadratic computational cost of $O(N^2 \cdot d)$ that bottlenecks memory and throughput, the FFU leverages the Fast Fourier Transform (FFT) to achieve $O(N \log N \cdot d)$ complexity.

    Eq.~\ref{eq3} defines the core FFU operation. The input sequence $\mathbf{X} \in \mathbb{R}^{N \times d}$ is reshaped into a 2D representation $\mathbf{X}^R \in \mathbb{R}^{H \times W \times d}$, projected into the frequency domain via a 2D Discrete Fourier Transform (DFT), $\mathcal{F}$, and modulated through element-wise multiplication with a learnable, input-independent spectral filter $\mathbf{W}_{freq}$. The modulated features are then restored to the spatial domain using an Inverse DFT ($\mathcal{F}^{-1}$). 
    Furthermore, we introduce a Global and Local Masking (GLM) mechanism  on $W_{freq}$ to suppress shortcut learning. local-masking (LM) mechanism prevents tokens from attending to their own positions.
    Considering the global modeling strength of FFU and ViT, all tokens in the same sequence may share similar information, we propose to mask tokens globally and randomly. 
    Specifically, by the convolution theorem, frequency-domain multiplication corresponds to spatial circular convolution:
    \begin{equation}\label{eq3}
    \mathbf{Y} = \text{FFU}(\mathbf{X})  = \mathcal{F}^{-1}(\mathcal{F}(\mathbf{X}^R) \odot \mathbf{W}_{freq}) = \mathbf{X}^R \circledast k
    \end{equation}
    where $k = \mathcal{F}^{-1}(\mathbf{W}_{freq})$ represents the spatial kernel, $\odot$ denotes the element-wise multiplication, $\circledast$ indicates the convolution operation. As shown in Fig.~\ref{fig2_new}(c), we set the origin of $k$ to zero (Local Masking) to prevent self-reproduction and apply Dropout on $k$ (Global Masking) to impair the copy from other tokens. The constrained spectral filter $\mathbf{W}^M_{freq}$ is then derived as:
    \begin{equation}\label{eq4}
    \mathbf{W}^{M}_{freq} = \mathcal{F}(k \odot M_{local} \odot M_{global})
    \end{equation}
    The final formulated FFU is expressed as:
    \begin{equation}\label{eq5}
    \text{FFU}(\mathbf{X}) = \mathcal{F}^{-1}(\mathcal{F}(\mathbf{X}^R) \odot \mathbf{W}^{M}_{freq})
    \end{equation}

		\begin{table*}[t]
		\centering
		\caption{Comparison between our method and currently state-of-the-art methods on four datasets, MVTec-AD \citep{bergmann2019mvtec}, ViSA \citep{zou2022spot}, Universal Medical \citep{he2024diffusion} and Real-IAD \citep{wang2024real}. Bold values indicate the best, and underlined values indicates the second best.  (\%)}
		\label{table1}
		\setlength{\tabcolsep}{3pt}
		\resizebox{1.0\textwidth}{!}{
			\begin{tabular}{c|ccccccc|ccccccccc}
				\toprule
				\multirow{3}{*}{Method} & \multicolumn{7}{c}{MVTec-AD} & \multicolumn{7}{c}{ViSA} \\
				\cmidrule{2-8} \cmidrule{9-15}
				& \multicolumn{3}{c}{Image-Level} & \multicolumn{4}{c|}{Pixel-Level} & \multicolumn{3}{c}{Image-Level} & \multicolumn{4}{c}{Pixel-Level} \\
				\cmidrule{2-5} \cmidrule{6-8} \cmidrule{9-12} \cmidrule{13-15}
				& AUC & AP & F1 & AUC & AP & F1 & AUPRO & AUC & AP & F1 & AUC & AP & F1 & AUPRO \\
				\midrule
				UniAD    & 96.5 & 98.8 & 96.2 & 96.8 & 43.4 & 49.5 & 90.7 & 88.8 & 90.8 & 85.8 & 98.3 & 33.7 & 39.0 & 85.5 \\
				Recontrast    & 98.3 & 99.4 & 97.6 & 97.1 & 60.2 & 61.5 & 93.2 & 95.5 & 96.4 & 92.0 & 98.5 & 47.9 & 50.6 & 91.9 \\
                ViTAD   & 98.3 & 99.4 & 97.3 & 97.7 & 55.3 & 58.7 & 91.4 & 90.5 & 91.7 & 86.3 & 98.2 & 36.6 & 41.1 & 85.1 \\
				DiAD      & 97.2 & 99.0 & 96.5 & 96.8 & 52.6 & 55.5 & 90.7 & 86.8 & 88.3 & 85.1 & 96.0 & 26.1 & 33.0 & 75.2 \\
				MambaAD        & 98.6 & 99.6 & 97.8 & 97.7 & 56.3 & 59.2 & 93.1 & 94.3 & 94.5 & 89.4 & 98.5 & 39.4 & 44.0 & 91.0 \\
				Dinomaly      & \underline{99.6} & \underline{99.8} & \underline{99.0} & \underline{98.4} & \underline{69.3} & \underline{69.2} & \underline{94.8} & \textbf{98.7} & \textbf{98.9} & \textbf{96.2} & \underline{98.7} & \underline{53.2} & \textbf{55.7} & \textbf{94.5} \\
				\textbf{Ours} & \textbf{99.8} & \textbf{99.9} & \textbf{99.4} & \textbf{98.5} & \textbf{71.5} & \textbf{70.5} & \textbf{95.6} & \textbf{98.7} & \textbf{98.9} & \underline{95.9} & \textbf{99.0} & \textbf{53.9} & \underline{55.0} & \textbf{94.5} \\
				\midrule[1.5pt]
				\multirow{3}{*}{Method} & \multicolumn{7}{c}{Universal Medical} & \multicolumn{7}{c}{Real-IAD} \\
				\cmidrule{2-8} \cmidrule{9-15}
				& \multicolumn{3}{c}{Image-Level} & \multicolumn{4}{c|}{Pixel-Level} & \multicolumn{3}{c}{Image-Level} & \multicolumn{4}{c}{Pixel-Level} \\
				\cmidrule{2-5} \cmidrule{6-8} \cmidrule{9-12} \cmidrule{13-15}
				& AUC & AP & F1 & AUC & AP & F1 & AUPRO & AUC & AP & F1 & AUC & AP & F1 & AUPRO \\
				\midrule
				UniAD           & 78.5 & 75.2 & 76.6 & 96.4 & 37.6 & 40.2 & 85.0 & 83.0 & 80.9 & 74.3 & 97.3 & 21.1 & 29.2 & 86.7 \\
				Reconstrast   & 80.1 & 79.7 & 80.9 & 96.3 & 42.3 & 43.8 & 85.2 & 86.4 & 84.2 & 77.4 & 97.8 & 31.6 & 38.2 & 91.8 \\
				ViTAD        & 82.2 & 81.0 & 80.1 & \textbf{97.1} & 49.9 & 49.6 & 86.1 & 82.3 & 79.4 & 73.4 & 96.9 & 26.7 & 34.9 & 84.9 \\
                DiAD          & \underline{85.1} & \underline{84.5} & 81.2 & 95.9 & 38.0 & 35.6 & 85.4 & 75.6 & 66.4 & 69.9 & 88.0 & 2.9 & 7.1 & 58.1 \\
				MambaAD         & 83.7 & 80.1 & \underline{82.0} & \underline{96.9} & 45.4 & 47.3 & \textbf{87.5} & 86.3 & 84.6 & 77.0 & 98.5 & 33.0 & 38.7 & 90.5 \\
				Dinomaly      & 84.9 & 84.1 & 81.0 & 96.8 & \underline{51.7} & \underline{52.1} & 85.5 & \underline{89.3} & \underline{86.8} & \underline{80.2} & \underline{98.8} & \underline{42.8} & \underline{47.1} & \underline{93.9} \\
				\textbf{Ours}                    & \textbf{87.6} & \textbf{87.0} & \textbf{82.5} & \textbf{97.1} & \textbf{56.0} & \textbf{55.6} & \underline{87.3} & \textbf{90.8} & \textbf{88.2} & \textbf{81.5} & \textbf{98.9} & \textbf{46.6} & \textbf{49.7} & \textbf{95.0} \\
				\bottomrule
		\end{tabular}}
        \label{evaluation}
		\vspace{-0.2in}
	\end{table*}

	\section{Experiments}
	\label{sec: Experiments}
	\subsection{Experimental Setting}
	\textbf{Datasets} We evaluated our method on four well-established datasets, including three for industrial scenarios: \textbf{MVTec-AD}~\citep{bergmann2019mvtec}, \textbf{ViSA}~\citep{zou2022spot}, \textbf{Real-IAD}~\citep{wang2024real}, and one for medical scenario: \textbf{Universal Medical}~\citep{he2024diffusion, he2024mambaad, zhang2023exploring, zhang2024invad}. \textbf{MVTec-AD} comprises 15 object categories, with 3,629 normal images in the training set and 498 normal images alongside 1,982 anomalous samples in the test set. \textbf{ViSA} contains 12 categories, providing 8,659 normal training images and a test set with 962 normal images and 1,200 anomalous cases. \textbf{Real-IAD}, the largest industrial benchmark, includes 30 diverse objects, utilizing 36,645 normal images for training and 63,256 normal images combined with 51,329 anomalous instances for testing. \textbf{Universal Medical} consists of 13,339 normal cases for training, with a test set containing 2,514 normal cases and 4,499 abnormal cases. This dataset spans three medical imaging modalities: Brain MRI, Liver CT, and Retinal CT scans. All datasets include both image-level and pixel-level labels.
	\textbf{Metrics} For evaluation metrics, we followed protocols from~\citep{zhang2023exploring, he2024mambaad, guo2025dinomaly}, adopting four metrics: Area Under the Receiver Operating Curve (AUC), F1-max score, average precision (AP) for both image-level detection and pixel-level localization tasks, and Area Under the Per-Region-Overlap (AUPRO) to further evaluate the localization task.
	\textbf{Implementation details} In our experiments, the encoder is initialized with DINO pre-trained weights~\citep{darcet2023vision} and kept frozen during training. Images are resized to 448×448 and then center-cropped to 392×392. We use a stable variant of the AdamW optimizer~\citep{wortsman2023stable} incorporating the AMSGrad algorithm~\citep{reddi2019convergence}, with a batch size of 32. Training iterations are set to 20,000 for all four datasets. The learning rate is initialized to 2e-3 and gradually reduced to 2e-4 via a Cosine Annealing schedule with a warm-start scheme~\citep{loshchilov2016sgdr}. The model is optimized using the global hard-mining loss~\citep{guo2025dinomaly}. In LRNB, we set the noise rate to 0.4 and the $r$ to 4. Similar to Dinomaly, set the number of LRNB layers to 1 for MVTec-AD and ViSA datasets, and 2 for more complex Universal Medical and Real-IAD datasets.

	\subsection{Evaluations}
\begin{figure*}{  
    \centering
    \includegraphics[width=0.98 \linewidth]{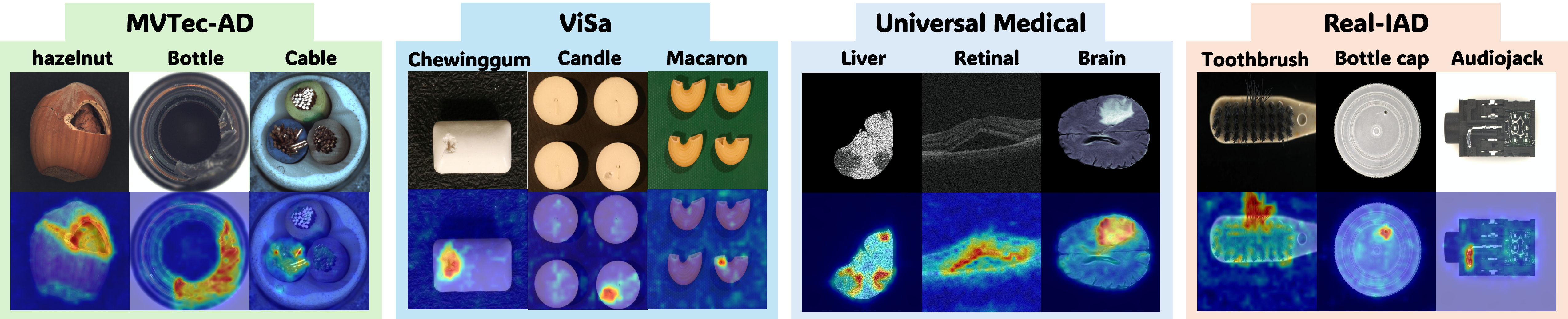}	

    \caption{Qualitative results of our Shortcutbreaker on the MVTec-AD, VisA, Universal Medical, and Real-IAD datasets. The first row displays the input images with their ground truth, while the second row presents the corresponding anomaly maps.
    }
    \vspace{-0.1in}
    \label{fig5_new}}
\end{figure*}

	\textbf{Comparison with SOTA MUAD methods}
	We evaluate our method against seven state-of-the-art (SOTA) MUAD methods—UniAD \citep{you2022unified}, DiAD \citep{he2024diffusion}, ViTAD~\citep{zhang2023exploring}, Recontrast~\citep{guo2024recontrast}, and Dinomaly~\citep{guo2025dinomaly}—across four benchmark datasets: MVTec-AD \citep{bergmann2019mvtec}, ViSA \citep{zou2022spot}, Universal Medical \citep{he2024diffusion}, and Real-IAD \citep{wang2024real}.
	Performance is quantified using both image-level metrics (I-AUC, I-AP, I-F1) and pixel-level metrics (P-AUC, P-AP, P-F1, P-AUPRO), where higher values indicate better detection capability.
	Experimental results are presented in Table~\ref{table1}, where our method outperforms comparative methods across all datasets in most metrics (The qualitative results obtained by our method can be seen in Fig.~\ref{fig5_new}).
	On the widely adopted MVTec-AD, our method achieves SOTA overall performance, with the highest image-level metrics of \textbf{99.8/99.9/99.4}, and pixel-level metrics of \textbf{98.5}/\textbf{71.5}/\textbf{70.5}/\textbf{95.6}. On ViSA, our method consistently achieves the superior overall performance, i.e., image-level metrics of \textbf{98.7}/\textbf{98.9}/\underline{95.9} and pixel-level metrics of \textbf{99.0}/\textbf{53.9}/\underline{55.0}/\textbf{94.5}.
	These results demonstrate that image-level performance on these two datasets is nearly 100\%, making performance improvements subtle.
	On Universal Medical, our method attains the highest image-level metrics of \textbf{87.6/87.0/82.5}, surpassing prior SOTAs by a large margin of 2.7/2.9/1.5, and achieves best or second-best pixel-level metrics of \textbf{97.1}/\textbf{56.0/55.6}/\underline{87.3}.
	On Real-IAD, our method produces a new SOTA result, with image-level and pixel-level performance of \textbf{90.8/88.2/81.5} and \textbf{98.9/46.6/49.7/95.0}, outperforming previous SOTAs by 1.5/1.4/1.3 and 0.1/3.4/2.6/1.1. These results demonstrate strong generalization to complex medical and diverse real-world industrial scenarios.

    \begin{table*}[t]
    \caption{Ablations studies of component contributions on Real-IAD and MvTec dataset, including LRNB: low-rank noisy bottleneck, FFB: frequency filtering block, GF: global filtering and GLM: global-local masking. (\%) }
    \centering
\begin{tabular}{ccc|cc|cc}
\toprule
\multirow{2}{*}{LRNB} & \multicolumn{2}{c|}{FFB} & \multicolumn{2}{c|}{Real-IAD}        & \multicolumn{2}{c}{MvTec-AD}         \\ \cmidrule{2-7} 
                      & GF         & GLM         & I-AUC/AP/F1    & P-/AUC/AP/F1/AUPRO  & I-AUC/AP/F1    & P-/AUC/AP/F1/AUPRO  \\ \midrule
\multicolumn{3}{c|}{basemodel}                   & 84.7/81.3/75.8 & 97.6/32.7/39.8/91.4 & 98.6/99.1/97.8 & 97.3/61.7/62.9/93.0 \\
$\checkmark$                     &           &            & 88.7/86.6/79.5 & 98.5/40.7/46.4/93.5 & 99.7/99.8/99.4 & 98.4/70.1/69.8/94.9 \\
                     & $\checkmark$          &            & 84.8/81.4/75.8 & 97.7/32.8/39.8/91.4 & 98.8/99.3/97.8 & 97.2/61.3/63.1/92.8 \\
                     & $\checkmark$          & $\checkmark$           & 85.3/81.9/76.0 & 97.7/33.5/40.4/91.5 & 99.1/99.5/98.3 & 97.3/62.1/63.4/93.0 \\
$\checkmark$                     & $\checkmark$          &            & 89.6/86.5/80.4 & 98.8/46.3/48.9/94.4 & 99.6/99.8/99.1 & 98.2/69.7/69.1/94.8 \\ 
$\checkmark$                     & $\checkmark$          & $\checkmark$           & \textbf{90.8}/\textbf{88.2}/\textbf{81.5} & \textbf{98.9}/\textbf{46.6}/\textbf{49.7}/\textbf{95.0} & \textbf{99.8}/\textbf{99.9}/\textbf{99.4} & \textbf{98.5}/\textbf{71.5}/\textbf{70.5}/\textbf{95.6} \\ 
\bottomrule
\label{table2_new}
\end{tabular}
\vspace{-0.25in}
\end{table*}

\begin{table}
    \caption{Effect of dual injection in LRNB on Real-IAD, G1 and G2 denotes the injection of LRNB before group 1 and group 2 decoder.  (\%)}
    \centering
			\resizebox{0.48\textwidth}{!}{
			\setlength{\tabcolsep}{3pt}
        \begin{tabular}{l|ccc|cccc}
        \toprule
        LRNB & I-AUC/AP/F1 & P-/AUC/AP/F1/AUPRO \\ \midrule
        None            & 85.3/81.9/76.9            & 97.7/33.4/30.4/91.5                   \\
        G1        &  90.2/87.5/81.1          &  98.9/45.2/48.5/94.8                  \\
        G2       &  88.7/86.0/79.3           &  98.0/45.6/49/1/94.0                  \\
        G1+G2        &          \textbf{90.8/88.2/81.5}   &  \textbf{98.9/46.6/49.7/95.0 }                 \\
        \bottomrule
    \end{tabular}%
    }
    \label{new_table_1}
    \vspace{-0.1in}
\end{table}

\begin{table}[t]
    \caption{Effect of global and local masking in FFB on Real-IAD, GM: global masking, LM: local masking and GLM: global-local masking.  (\%)}
    \centering
			\resizebox{0.48\textwidth}{!}{
			\setlength{\tabcolsep}{3pt}
        \begin{tabular}{l|ccc|cccc}
        \toprule
        FFB & I-AUC/AP/F1 & P-/AUC/AP/F1/AUPRO \\ \midrule
        None            & 89.6/86.5/80.4           & 98.8/46.3/48.9/94.4                   \\
        
        GM        &   89.8/86.9/80.5          & 98.8/46.3/49.2/94.7               \\
        LM       &  90.0/86.8/80.6           &  99.0/46.4/49.2/94.7                 \\
        GLM        &         \textbf{ 90.8/88.2/81.5 }  &  \textbf{98.9/46.6/49.7/95.0}                  \\
        \bottomrule
    \end{tabular}%
    }
    \vspace{-0.2in}
    \label{new_table_2}
\end{table}

	\subsection{Ablation Studies}
	\noindent \textbf{Overall Ablation} To explore the contributions of each component in our method, including the low-rank noisy bottleneck (LRNB), global filtering (GF), and global-local masking (GLM) operations, we conduct ablation experiments as shown in Table~\ref{table2_new}.
	The baseline is constructed following Dinomaly \citep{guo2025dinomaly} and ViTAD \citep{zhang2023exploring}, which construct a baseline model with a DINO-pretrained ViT encoder and a learnable softmax attention-based ViT decoder. The effectiveness of the baseline has been proven in industrial scenarios.
	The results of the ablation experiments demonstrate that the proposed LRNB, GF and GLM modules all contribute to performance improvement, with their combinations further enhancing performance.
	Compared to the baseline model, using any single module alone achieves better performance in terms of the key indicators I-AUC and P-AUC.
	Among them, FFB (GF with GLM) brings moderate improvements, while LRNB shows the most significant enhancement—it notably boosts image-level and pixel-level performances with respective large margins of 5.1/6.9/5.7 and 2.3/13.9/9.9/3.6 on Real-IAD dataset and 1.1/0.7/1.6, and 1.1/8.4/6.9/1.9 on MvTec-AD dataset. 
    These results demonstrate that the LRNB module serves as a core foundation in our method.
	When replacing self-attention with sole GF, the overall improvements on two datasets are subtle, however the computational complexity is reduced from $O(N^2 \cdot d)$ to $O(N \log N \cdot d)$.
    Further combination with GLM yields even better results.
	The final integration of all three modules (LRNB + GF + GLM) delivers the best overall performance, achieving the highest image-level and pixel-level metrics on both datasets—confirming that their collaborative effect effectively enhances the model's overall performance.

    Table~\ref{new_table_1} illustrates the impact of dual injection within the LRNB framework. The results demonstrate that injecting LRNB before either the Group 1 (G1) or Group 2 (G2) decoder yields significant performance improvements. Notably, G1 injection serves as the primary contributor, achieving superior accuracy compared to G2. We attribute this to the fact that G1 injection influences all subsequent decoder blocks, whereas G2 only affects the final four blocks. Furthermore, the combination of G1 and G2 surpasses single-injection strategies at both the image and pixel levels by further enlarging the score gap between normal and abnormal cases. The corresponding training loss and anomaly score plots in Fig.~\ref{new_table_1} corroborate these findings.
    Table~\ref{new_table_2} evaluates Global-Local Masking (GLM). While GM or LM alone yields marginal gains, combining them significantly boosts performance. This confirms our observation that ViT features suffer from information leakage via both self-identity and long-range token correlations, necessitating a dual-masking approach. The GLM prevents the shortcuts globally and locally, thus enhancing the anomaly detection ability.
	
\begin{figure}  
    \centering
    \includegraphics[width=8cm, height=4cm]{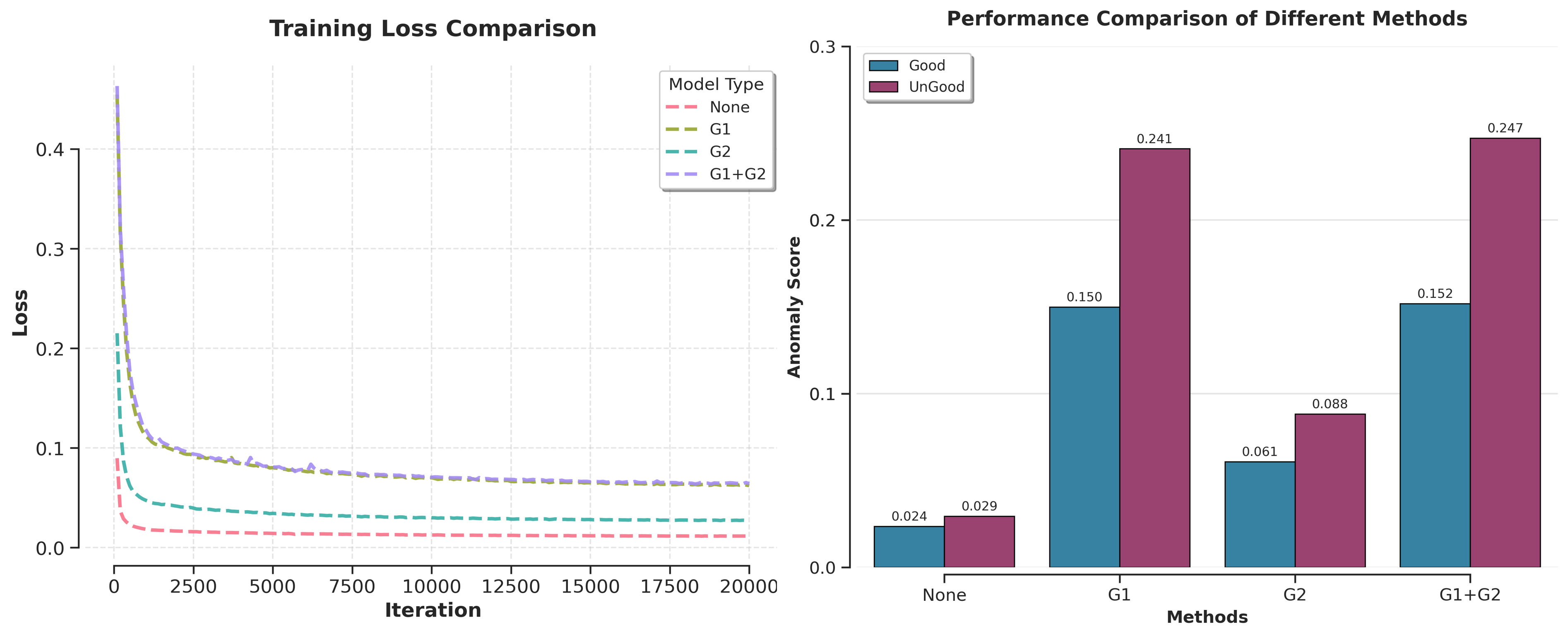}	
    \caption{The corresponding plots of training loss (left part) and the averaged anomaly scores (right part) of the methods in Table~\ref{new_table_1}.
    }
    \vspace{-0.25in}
    \label{new_fig1}
\end{figure}

		\begin{table*}[htbp]
		\centering
		\begin{minipage}[t]{0.48\linewidth}
            \caption{Performance comparison of different bottlenecks (BNs) on Real-IAD. G1 denotes the bottleneck is only injected before group 1 decoder (\%)}

			\centering
			\resizebox{1.0\textwidth}{!}{
			\setlength{\tabcolsep}{3pt}
				\begin{tabular}{l|ccc|cccc}
					\toprule
                    Bollteneck           & I-AUC/AP/F1 & P-/AUC/AP/F1/AUPRO \\ \midrule
                    None                 &  84.7/81.3/75.8           &  97.6/32.7/39.8/91.4                   \\
                    FJ (G1)                  & 87.6/84.5/78.3            & 98.3/40.1/44.8/92.3                   \\
                    NBD (G1)                 &   88.5/86.0/79.2          & 98.5/43.3/47.1/93.5                   \\
                    LRNB (G1)                 &   90.2/87.5/81.1          &   98.9/45.2/48.5/94.8                 \\
                    LRNB (G1+G2) & \textbf{90.8/88.2/81.5}            &  \textbf{98.9/46.6/49.7/95.0}                  \\
                    \bottomrule
				\end{tabular}%
			}
			\label{table3_new}
		\end{minipage}
		\hfill
		\begin{minipage}[t]{0.48\linewidth}
            \caption{Performance comparison of different modules in decoder on Real-IAD. (\%)}
			\centering
			\resizebox{1.0\textwidth}{!}{
			\setlength{\tabcolsep}{3pt}
				\begin{tabular}{l|ccc|cccc}
				\toprule
                Decoder Blocks & I-AUC/AP/F1 & P-/AUC/AP/F1/AUPRO \\ \midrule
                ViT            & 88.7/86.6/79.5            & 98.5/40.7/46.4/93.5                   \\
                CNN+ViT        &  89.0/86.1/79.5           &  98.8/45.5/48.7/94.4                  \\
                NMA+ViT        &  90.4/88.0/81.2           &  98.9/45.4/48.9/94.6                  \\
                LSA        &          89.3/87.5/79.9   &  98.6/43.4/47.9/93.9                  \\
                FFB (ours)           &            \textbf{90.8/88.2/81.5}            &  \textbf{98.9/46.6/49.7/95.0}                          \\
				\bottomrule
			\end{tabular}%
			}
			\label{table4_new}
		\end{minipage}
        \vspace{-0.2in}
	\end{table*}

	\noindent \textbf{Comparison between LRNB and previous noisy bottlenecks}
	Previous methods also proposed some noisy operations in the bottleneck to avoid identity shortcuts, such as feature jittering (FJ) in UniAD \citep{you2022unified} and noisy dropout bottleneck (NDB) in Dinomaly \citep{guo2025dinomaly}. To further validate the advantage of the proposed LRNB, we conduct a comparison in Table \ref{table3_new}.
	The results in this table demonstrate that FJ experiences a performance drop and NDB yields slight improvements compared to models without such noise operations, while our LRNB and LRNB-D achieve a significant enhancement.
\begin{table}[h]
\begin{minipage}[t]{0.98\linewidth}
\caption{Comparison of computational cost of different methods on Real-IAD, mAD is the average value of all the metrics. Shortcutbreaker-s is based on the encoder of ViT-small model, Shortcutbreaker-s* is based on  ViT-small model and trained with 224x224 images.}
\centering
\resizebox{1.0\textwidth}{!}{
\setlength{\tabcolsep}{3pt}
\begin{tabular}{l|ccc}
\toprule
Method          & Params(M) & FLOPS(G) & mAD  (\%)         \\ \midrule
UniAD           & \textbf{24.5}      & \textbf{3.6}      & 67.5          \\
DiAD            & 1331.3    & 451.5    & 52.6          \\
MambaAD         & \underline{25.7}      & 8.3      & 72.7          \\
Dinomaly        & 132.8     & 104.7    & 77.0          \\
\midrule
\textbf{Shortcutbreaker} & 109.8     & 86.5     & \textbf{78.7} \\ 
\textbf{Shortcutbreaker-s} & 27.6     & 21.8     & \underline{77.5} \\ 
\textbf{Shortcutbreaker-s*} & 27.6     & \underline{7.19}     & 73.3 \\ 

\bottomrule
\end{tabular}}
\end{minipage}
\label{tab:computational_complexity}
\vspace{-0.1in}
\end{table}

\begin{figure}  
    \centering
    \includegraphics[width=8cm, height=4cm]{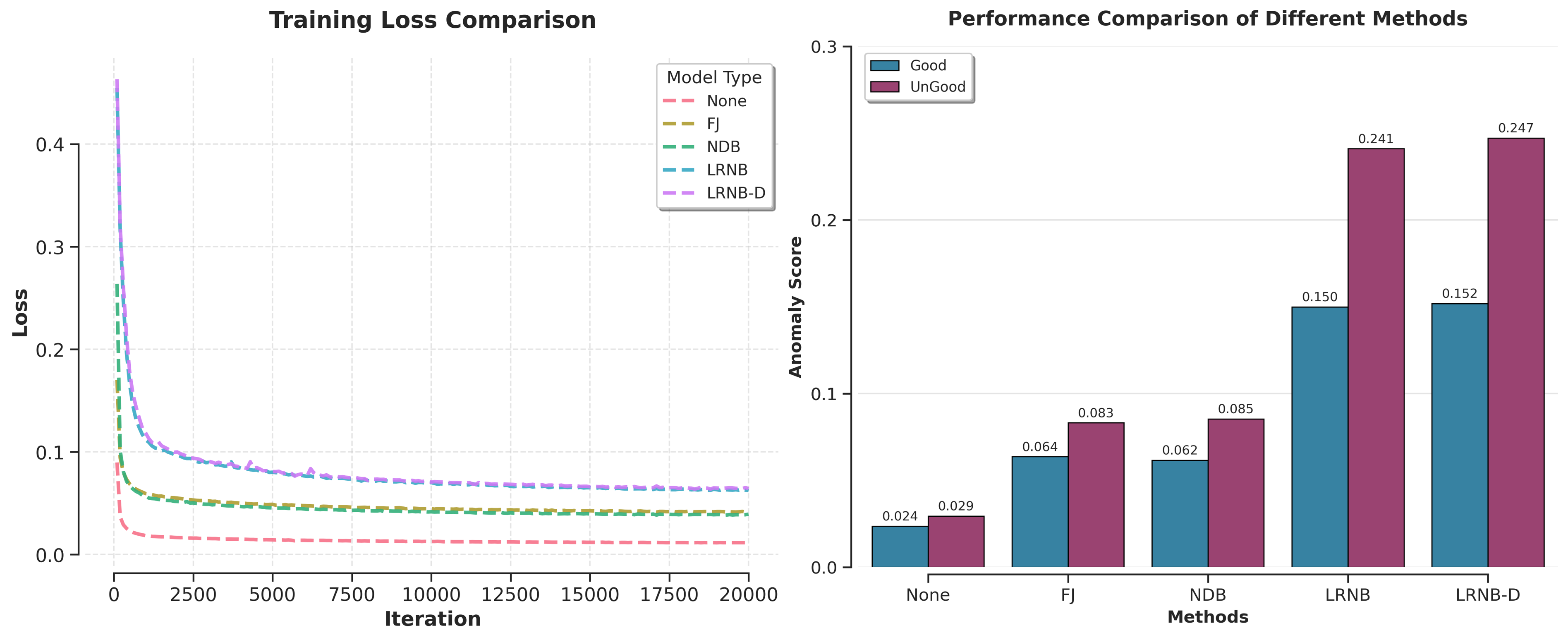}	
    \caption{The plots of training loss (left part) and the averaged anomaly scores (right part) of different bottlenecks on Real-IAD.
    }
    \vspace{-0.25in}
    \label{fig6_new}
\end{figure}

	To explore the capacity of these operations in addressing the "identity mapping" issue, we further examine the corresponding training loss and average anomaly score.
	As shown in Fig.~\ref{fig6_new}, models without a bottleneck achieve near-zero training loss and obtain extremely close anomaly scores between normal and abnormal cases (0.024 vs. 0.029). These results indicate that these models suffer from identity mapping behavior where inputs are directly replicated in outputs.
	Introducing FJ or NDB noise operations partially alleviates this issue, which we attribute to enhanced noise robustness from exposure to diverse industrial patterns.
	Our LRNB and LRNB-D effectively resolve identity mapping: while moderately increasing the score of normal data, it substantially enhances the score of abnormal cases.  This strategic trade-off widens the discrimination margin between normal and abnormal samples, ultimately improving anomaly detection performance.
	\noindent \textbf{Comparison between FFB and previous modules in decoder}
	Prior works have introduced specialized decoder modules to mitigate identity mapping. For instance, \citep{lu2024anomaly} attributes this issue to encoder-decoder homogeneity and advocates for heterogeneous structures, such as using CNN blocks to reconstruct ViT encoder outputs. Similarly, \citep{you2022unified} introduced neighbor-masking attention (NMA) to prevent information leakage, while \citep{guo2025dinomaly} utilized linear self-attention (LSA) to "un-focus" features for robust reconstruction. As shown in Table \ref{table4_new}, replacing the ViT baseline with a hybrid ViT+CNN structure improves performance, with LSA providing further gains. Additionally, NMA proves effective in the ViT decoder, yielding improvements of 0.7/1.4/1.7 in image-level and 0.1/4.7/2.5/1.0 in pixel-level metrics over pure ViT with negligible computational overhead. Ultimately, our FFB outperforms all existing methods across all metrics, which we attribute to its superior adaptability to ViT-extracted tokens.

    \noindent \textbf{Comparison of computational complexity} Table~\ref{tab:computational_complexity} presents the trade-off between computational complexity and mAD on Real-IAD dataset. where our Shortcutbreaker framework demonstrates clear superiority over state-of-the-art methods. Notably, the primary Shortcutbreaker variant achieves a peak performance of 78.7 mAD, surpassing the second-best Dinomaly while reducing computational costs by 17.4\%. Furthermore, Shortcutbreaker-S and Shortcutbreaker-S*offer exceptional scalability; the former maintains a superior 77.5 mAD with only 27.6M parameters, while the latter delivers a highly efficient 7.19G FLOPs for real-time applications, even better than the light-weight structure of MambaAD \cite{he2024mambaad}. 


		\section{Conclusion}
    We propose ShortcutBreaker, a novel feature-reconstruction framework designed to mitigate identity shortcuts in the MUAD setting. It introduces two core innovations: a low-rank noisy bottleneck (LRNB) and a frequency-filtering block (FFB), which prevent shortcut learning within the bottleneck and decoder components. Extensive experiments across four well-established benchmarks confirm the efficacy of these modules, demonstrating consistent superiority over state-of-the-art methods, particularly in diverse scenarios. Furthermore, its low GFLOPs underline its potential for real-time applications.	

    \section*{Impact Statement}
     This paper presents work whose goal is to advance the field of  Machine Learning. There are many potential societal consequences of our work, none of which we feel must be specifically highlighted here.

	
	\bibliography{example_paper}
	\bibliographystyle{icml2026}
	\newpage
	\appendix
	\onecolumn
	\section{Appendix}
	
	\subsection{Proofs}
	\label{sec: proof}
	
	In this section, we will provide the proof of the first property of low-rank matrix decomposition under non-linear transformation, the analysis of rank constraints of the proposition 1 in \cite{cai2024rethinking}. In our proofs, we assume the input feature matrix $x \in \mathbb{R}^d$  maintains full column rank. This assumption aligns with our empirical observations from practical computations using Singular Value Decomposition (SVD). 
	
	\paragraph{Analysis of rank constraints of the proposition 1 in \citep{cai2024rethinking}}
	\label{sec: flawed_inference} 
	Given the condition of linear transformation, Proposition 1 in  \citep{cai2024rethinking} establishes that for $k \geq \frac{d}{2}$, the $W_1W_2=I_{d \times d}$ with $W_1 \in \mathbb{R}^{d \times k}$ and $W_2 \in \mathbb{R}^{k \times d}$ admits at least one solution. Consequently, in turns, when  no solutions exist for this equation, $k < \frac{d}{2}$, thereby inherently suppressing identity mapping.
	However, a critical paradox emerges in the transitional regime $d>k\geq\frac{d}{2}$: Proposition 1 guarantees solution existence, but the rank inequality  $r(W_1W_2)\leq min(r(W_1),r(W_2))\leq k< d$ directly conflicts with the full-rank requirement $I_{d \times d} = d$. 
	Therefore, this reasoning process is flawed.

	\begin{table}[h]
		\centering
		\caption{Effect of dual injection in LRNB on Real-IAD and MvTec-AD, G1 and G2 denotes the injection of LRNB before group 1 and group 2 decoder.}
		\begin{tabular}{l|cccc}
			\toprule
			\multirow{2}{*}{LRNB} & \multicolumn{2}{c}{Real-IAD}         & \multicolumn{2}{c}{MvTec-AD}            \\
			& I-AUC/AP/F1    & P-/AUC/AP/F1/AUPRO  & I-AUC/AP/F1    & P-/AUC/AP/F1/AUPRO  \\ \midrule
			None                  & 84.7/81.3/75.8 & 97.6/32.7/39.8/91.4 & 99.1/99.5/98.3 & 97.3/62.1/63.4/93.0 \\
			G1                    & 90.2/87.5/81.1 & 98.9/45.2/48.5/94.8 & 99.8/99.9/99.4 & 98.4/70.5/69.7/95.5 \\
			G2                    & 88.7/86.0/79.3 & 98.0/45.6/49/1/94.0 & 99.4/99.7/98.9 & 97.9/70.3/69.2/94.5 \\
			G1+G2                 & \textbf{90.8/88.2/81.5} & \textbf{98.9/46.6/49.7/95.0} & \textbf{99.8/99.9/99.4} & \textbf{98.5/71.5/70.5/95.6} \\ \bottomrule
		\end{tabular}
		\label{appendix:table1}
	\end{table}

	\begin{table}[h]
		\caption{Performance comparison of different bottlenecks on Real-IAD and MvTec-AD datasets. G1 denotes the bottleneck is only injected before group 1 decoder (\%)}
		
		\centering
		\begin{tabular}{l|cccc}
			\toprule
			\multirow{2}{*}{Bottleneck} & \multicolumn{2}{c}{Real-IAD}                           & \multicolumn{2}{c}{MvTec-AD}                           \\
			& I-AUC/AP/F1             & P-/AUC/AP/F1/AUPRO           & I-AUC/AP/F1             & P-/AUC/AP/F1/AUPRO           \\ \midrule
			None                        & 84.7/81.3/75.8          & 97.6/32.7/39.8/91.4          & 99.1/99.5/98.3          & 97.3/62.1/63.4/93.0          \\
			FJ   (G1)                       & 87.6/84.5/78.3          & 98.3/40.1/44.8/92.3          & 99.5/99.7/98.9          & 98.1/68.2/68.0/94.5          \\
			NBD  (G1)                       & 88.5/86.0/79.2          & 98.5/43.3/47.1/93.5          & 99.7/99.8/99.1          & 98.2/68.8/68.7/94.8          \\
			LRNB (G1)                       & 90.2/87.5/81.1          & 98.9/45.2/48.5/94.8          & 99.8/99.9/99.4          & 98.4/70.5/69.7/95.5          \\
			LRNB((G1+G2)        & \textbf{90.8/88.2/81.5} & \textbf{98.9/46.6/49.7/95.0} & \textbf{99.8/99.9/99.4} & \textbf{98.5/71.5/70.5/95.6} \\ \bottomrule
		\end{tabular}
		\label{appendix:table3}
	\end{table}
	
	\begin{table}[h]
		\caption{Performance comparison of different modules in decoder on Real-IAD and MvTec-AD dataset. (\%)}
		\centering
		\begin{tabular}{l|cccc}
			\toprule
			\multirow{2}{*}{Decoder} & \multicolumn{2}{c}{Real-IAD}         & \multicolumn{2}{c}{MvTec-AD}         \\
			& I-AUC/AP/F1    & P-/AUC/AP/F1/AUPRO  & I-AUC/AP/F1    & P-/AUC/AP/F1/AUPRO  \\ \midrule
			ViT                      & 88.7/86.6/79.5 & 98.5/40.7/46.4/93.5 & 99.7/99.8/99.4 & 98.4/70.1/69.8/94.9 \\
			CNN+ViT                  & 89.0/86.1/79.5 & 98.8/45.5/48.7/94.4 & 99.7/99.8/99.4 & 98.3/70.1/69.8/94.9 \\
			NMA+ViT                      & 90.4/88.0/81.2 & 98.9/45.4/48.9/94/6 & 99.7/99.9/99.4 & 98.4/71.2/69.9/95.1 \\
			LA                       & 89.3/87.5/79.9 & 98.6/43.4/47.9/93.9 & 99.2/99.7/98.9 & 97.5/62.5/64.8/92.6 \\
			FFB (Ours)                     & \textbf{90.8/88.2/81.5} & \textbf{98.9/46.6/49.7/95.0} & \textbf{99.8/99.9/99.4} & \textbf{98.5/71.5/70.5/95.6} \\ \bottomrule
		\end{tabular}
		\label{appendix:table2}
	\end{table}
	
	\subsection{Supplementary Ablation Studies}
	Due to the page limit of main text, we present more ablation studies in this section of Appendix.
	
	Table~\ref{appendix:table1} extends Table~\ref{new_table_1} to the MVTec-AD dataset, confirming the efficacy of the dual injection strategy. Observations remain consistent: although both G1 and G2 injections improve performance, G1 provides the most significant gain. Crucially, the G1 and G2 combination outperforms either standalone injection.
	
	Extending Table~\ref{table3_new} to MVTec-AD, Table~\ref{appendix:table2} validates our FFB. Compared to FJ and NDB, LRNB achieves substantial gains across both datasets. Specifically, incorporating dual injection strategy in LRNB further enhances results by effectively enlarging the score margin between normal and abnormal cases.
	
	Table~\ref{appendix:table2} extends Table~\ref{table4_new} to the MVTec-AD dataset, confirming the efficacy of our FFB. The results display that
	compared FJ and NDB, our LRNB significantly improve the performances on both datasets, further injection before G2 further improve the performance by widening the gap between normal and abnormal samples.

	\begin{table}[t]
		\centering
		\caption{The effect of dropout rate in LRNB. (\%)}
		\begin{tabular}{l|cccc}
			\toprule
			\multirow{2}{*}{LRNB nosie rate} & \multicolumn{2}{c}{Real-IAD}                                             & \multicolumn{2}{c}{MvTec-AD}                                             \\
			& \multicolumn{1}{c}{I-AUC/AP/F1} & \multicolumn{1}{c}{P-/AUC/AP/F1/AUPRO} & \multicolumn{1}{c}{I-AUC/AP/F1} & \multicolumn{1}{c}{P-/AUC/AP/F1/AUPRO} \\ \midrule
			0                                & 89.5/86.6/80.0                  & 98.9/45.2/48.1/94/1                    & 99.5/99.7/98.7                  & 98.1/67.2/67.0/94.2                    \\
			0.1                              & 89.9/87.1/80.7                  & 98.9/46.0/49.0/94.5                    & 99.6/99.7/99.1                  & 98.4/70.6/69.7/94.9                    \\
			0.2                              & 90.2/87.4/80.9                  & 98.9/46.8/49.2/94.7                    & 99.7/99.8/99.2                  & 98.5/71.3/70.2/95.3                    \\
			0.3                              & 90.3/87.5/81.0                  & 98.9/46.0/49.2/94.8                    & 99.7/99.9/99.4                  & 98.5/71.2/70.2/95.4                    \\
			0.4                              & \textbf{90.8/88.2/81.5}         & \textbf{98.9/46.6/49.7/95.0}           & \textbf{99.8/99.9/99.4}         & \textbf{98.5/71.5/70.5/95.6}           \\
			0.5                              & 90.6/88.0/81.5                  & 99.0/45.9/49.1/94.9                    & 99.7/99.9/99.4                  & 98.5/71.5/70.5/95.7                    \\ \bottomrule
		\end{tabular}
		\label{appendix:table4}
	\end{table}
	
	\begin{table}[t]
		\centering
		\caption{The effect of layer number in LRNB. (\%)}
		\begin{tabular}{l|cccc}
			\toprule
			\multirow{2}{*}{Layer number $i$} & \multicolumn{2}{c}{Real-IAD}         & \multicolumn{2}{c}{MvTec-AD}         \\
			& I-AUC/AP/F1    & P-/AUC/AP/F1/AUPRO  & I-AUC/AP/F1    & P-/AUC/AP/F1/AUPRO  \\ \midrule
			1                             & 89.4/87.1/80.2 & 98.7/46.2/48.9/94.0 & \textbf{99.8/99.9/99.4} & \textbf{98.5/71.5/70.5/95.6} \\
			2                             & \textbf{90.8/88.2/81.5} & \textbf{98.9/46.6/49.7/95.0} & 99.7/99.9/99.4 & 98.1/67.8/66.9/94.5 \\
			3                             & 90.3/87.6/81.1 & 99.0/45.2/48.8/94.8 & 99.5/99.9/99.2 & 97.6/59.9/61.1/92.7 \\ \bottomrule
		\end{tabular}
	\end{table}
	\label{appendix:table5}
	
	\begin{table}[h]
		\centering
		\caption{The effect of low-rank ratio $r$ in LRNB. (\%)}
		\begin{tabular}{l|cccc}
			\hline
			\multirow{2}{*}{low-rank ratio $r$} & \multicolumn{2}{c}{Real-IAD}                                             & \multicolumn{2}{c}{MvTec-AD}                                             \\
			& \multicolumn{1}{c}{I-AUC/AP/F1} & \multicolumn{1}{c}{P-/AUC/AP/F1/AUPRO} & \multicolumn{1}{c}{I-AUC/AP/F1} & \multicolumn{1}{c}{P-/AUC/AP/F1/AUPRO} \\ \hline
			2                               & 89.9/87.2/80.5                  & 98.9/46.5/49.0/94.4                    & 99.7/99.8/99.3                  & 98.3/70.8/70.0/95.3                    \\
			4                               & \textbf{90.8/88.2/81.5 }                 & \textbf{98.9/46.6/49.7/95.0 }                   & \textbf{99.8/99.9/99.4}                  & \textbf{98.5/71.5/70.5/95.6}                    \\
			6                               & 89.7/86.6/80.3                  & 98.8/44.7/48.2/84.6                    & 99.7/99.8/99.3                  & 98.4/71.4/70.4/95.6                    \\
			8                               & 90.0/87.1/80.7                  & 99.0/45.5/48.7/94.8                    & 99.3/99.7/98.8                  & 98.0/66.0/66.6/94.1                    \\
			10                              & 90.0/87.2/80.5                  & 98.9/45.0/47.8/94.4                    & 99.8/99.9/99.4                  & 98.4/70.6/70.1/95.8                    \\ \hline
		\end{tabular}
		\label{appendix:table7}
	\end{table}

	\begin{table}[h]
		\centering
		\caption{The effect of noise operation in downsampling blcoks (DBs) and upsamling blocks (UBs) of LRNB. (\%)}
		\begin{tabular}{cc|cccc}
			\toprule
			\multirow{2}{*}{UB} & \multirow{2}{*}{DB} & \multicolumn{2}{c}{Real-IAD}         & \multicolumn{2}{c}{MvTec-AD}         \\
			&                     & I-AUC/AP/F1    & P-/AUC/AP/F1/AUPRO  & I-AUC/AP/F1    & P-/AUC/AP/F1/AUPRO  \\ \midrule
			& $\checkmark$                   & 90.3/87.6/81.0 & 99.1/48.3/50.2/94.8 & 99.7/99.8/99.2 & 98.4/70.9/70.2/95.0 \\
			$\checkmark$                   &                     & \textbf{90.8/88.2/81.5} & \textbf{98.9/46.6/49.7/95.0} & \textbf{99.8/99.9/99.4} & \textbf{98.5/71.5/70.5/95.6} \\
			$\checkmark$                   & $\checkmark$                   & 90.2/87.5/81.0 & 98.9/47.4/49.7/95/0 & 99.8/99.9/99.4 & 98.5/71.2/70.2/95.6 \\ \bottomrule
		\end{tabular}
		\label{appendix:table6}
	\end{table}

	\begin{table}[t]
		\centering
		\caption{The effect of noise ratio in global masking (dropout operation on frequency-spectral fitler $W_{freq}$) of FFU. (\%)}
		\begin{tabular}{l|cccc}
			\toprule
			\multicolumn{1}{c|}{\multirow{2}{*}{noise ratio in GM}} & \multicolumn{2}{c}{Real-IAD}                           & \multicolumn{2}{c}{MvTec-AD}         \\
			\multicolumn{1}{c|}{}                                   & I-AUC/AP/F1             & P-/AUC/AP/F1/AUPRO           & I-AUC/AP/F1    & P-/AUC/AP/F1/AUPRO  \\ \midrule
			0                                                       & 89.6/86.5/80.4          & 98.8/46.3/48.9/94.4          & 99.7/99.8/99.4 & 98.4/71.0/70.2/95.3 \\
			0.1                                                     & 90.2/87.2/81.1          & 98.8/46.0/49.3/94.8          & 99.7/99.9/99.4 & 98.5/71.3/70.5/95.4 \\
			0.2                                                     & 90.5/87.9/81.4          & 98.9/46.1/49.5/94.9          & 99.7/99.9/99.4 & 98.4/71.2/70.4/95.5 \\
			0.3                                                     & \textbf{90.8/88.2/81.5} & \textbf{98.9/46.6/49.7/95.0} & \textbf{99.8/99.9/99.4} & \textbf{98.5/71.5/70.5}/95.6 \\
			0.4                                                     & 90.3/87.6/81.0          & 98.8/46.9/49.9/94.8          & 99.8/99.9/99.4 & \textbf{98.5/71.5/70.5/95.7} \\
			0.5                                                     & 90.4/87.8/81.2          & 98.9/45.7/49.1/94.8          & 99.8/99.9/99.4 & 98.5/71.5/70.4/95.6 \\ \bottomrule
		\end{tabular}
		\label{appendix:table8}
	\end{table}
	

	\begin{table}[h]
		\centering
		\caption{The effect of mask size in local masking (center zero operation on frequency-spectral fitler $W_{freq}$) of FFU. (\%)}
		\begin{tabular}{l|cccc}
			\toprule
			\multicolumn{1}{c|}{\multirow{2}{*}{noise ratio in GM}} & \multicolumn{2}{c}{Real-IAD}                           & \multicolumn{2}{c}{MvTec-AD}         \\
			\multicolumn{1}{c|}{}                                   & I-AUC/AP/F1             & P-/AUC/AP/F1/AUPRO           & I-AUC/AP/F1    & P-/AUC/AP/F1/AUPRO  \\ \midrule
			None                                                  & 89.6/86.5/80.4                  & 98.8/46.3/48.9/94.4                    & 99.6/99.8/99.1                  & 98.2/69.7/69.1/94.8                    \\
			3                                                     & 90.3/87.5/80.9                  & 98.9/46.6/49/7/94.9                    & 99.7/99.8/99.4                  & 98.4/70.7/70.1/95.0                    \\
			5                                                     & \textbf{90.8/88.2/81.5}         & \textbf{98.9/46.6/49.7/95.0}           & \textbf{99.8/99.9/99.4}         & \textbf{98.5/71.5/70.5/95.6}           \\
			7                                                     & 90.3/87.5/81.0                  & 98.9/46.2/49.3/94.8                    & 99.7/99.8/99.4                  & 98.4/71.5/70.4/95.6                    \\
			9                                                     & 89.9/87.1/80.7                  & 99.0/46.5/49.2/94.5                    & 99.7/99.9/99.4                  & 98.4/71.8/70.6/95.8                    \\ \hline
		\end{tabular}
		\label{appendix:table9}
	\end{table}
	
	\subsection{Hyperparameter Analysis}
	In this section, we will present the experiments about the hyper-parameter selection and analysis, which are divided into two parts: LRNB related hyperparameter and FFU related hyperparameter.
	
	\textbf{LRNB:} First, we investigate the hyper-parameter settings for LRNB, specifically the impact of dropout. Integrated to perturb tokens and mitigate identity mapping, this mechanism improves performance across the 0.1–0.5 range compared to the baseline (dropout=0.0), as shown in Table~\ref{appendix:table4}, with 0.4 yielding the best results.Second, we explore the effect of different LRNB layer number ($i$) in Table~\ref{appendix:table5}. Observations indicate the optimal setting is $i=1$ for MVTec-AD and $i=2$ for Real-IAD. This difference is attributed to varying dataset complexity; the more diverse Real-IAD dataset requires stronger noise perturbation to suppress identity shortcuts. Following Dinomaly~\cite{guo2025dinomaly}, we set $i=1$ for MVTec-AD and VisA, and $i=2$ for Real-IAD and Universal Medical datasets. Third, we investigate the impact of dropout noise within the downsampling and upsampling blocks (DBs and UBs). As shown in Table~\ref{appendix:table6}, the best performance is achieved when dropout operations are embedded solely in the UBs. 
	Forth, Table~\ref{appendix:table7} evaluates the impact of varying the LRNB low-rank ratio $r$, where a lower $r$ indicates more intense token perturbation. Optimal performance is achieved at $r=4$, suggesting that excessive perturbation can hinder the reconstruction of normal samples. Consequently, a balanced $r$ is essential to maximize the discriminative gap between normal and abnormal samples while preserving the reconstruction integrity of normal patterns.

	\textbf{FFU:} In this part, we explore the hyper-parameter setting for FFU.First, we investigate the impact of the Global Masking (GM) ratio (dropout on the spectral filter $W_{freq}$) in Table~\ref{appendix:table8}. The results demonstrate that GM consistently enhances performance, confirming its robustness, with an optimal noise ratio of 0.3. Second, we explore the influence of the mask size of Local Masking (LM)  in Table~\ref{appendix:table9}. The results indicate that the LM mechanism consistently improves performance compared to the baseline without local masking. The highest accuracy is achieved with a mask size of 5 across both datasets, further validating the effectiveness of local frequency-domain perturbation.
	
	\subsection{Detailed Results}
\begin{figure*}[t]
	\centering
	\includegraphics[width=0.95 \linewidth]{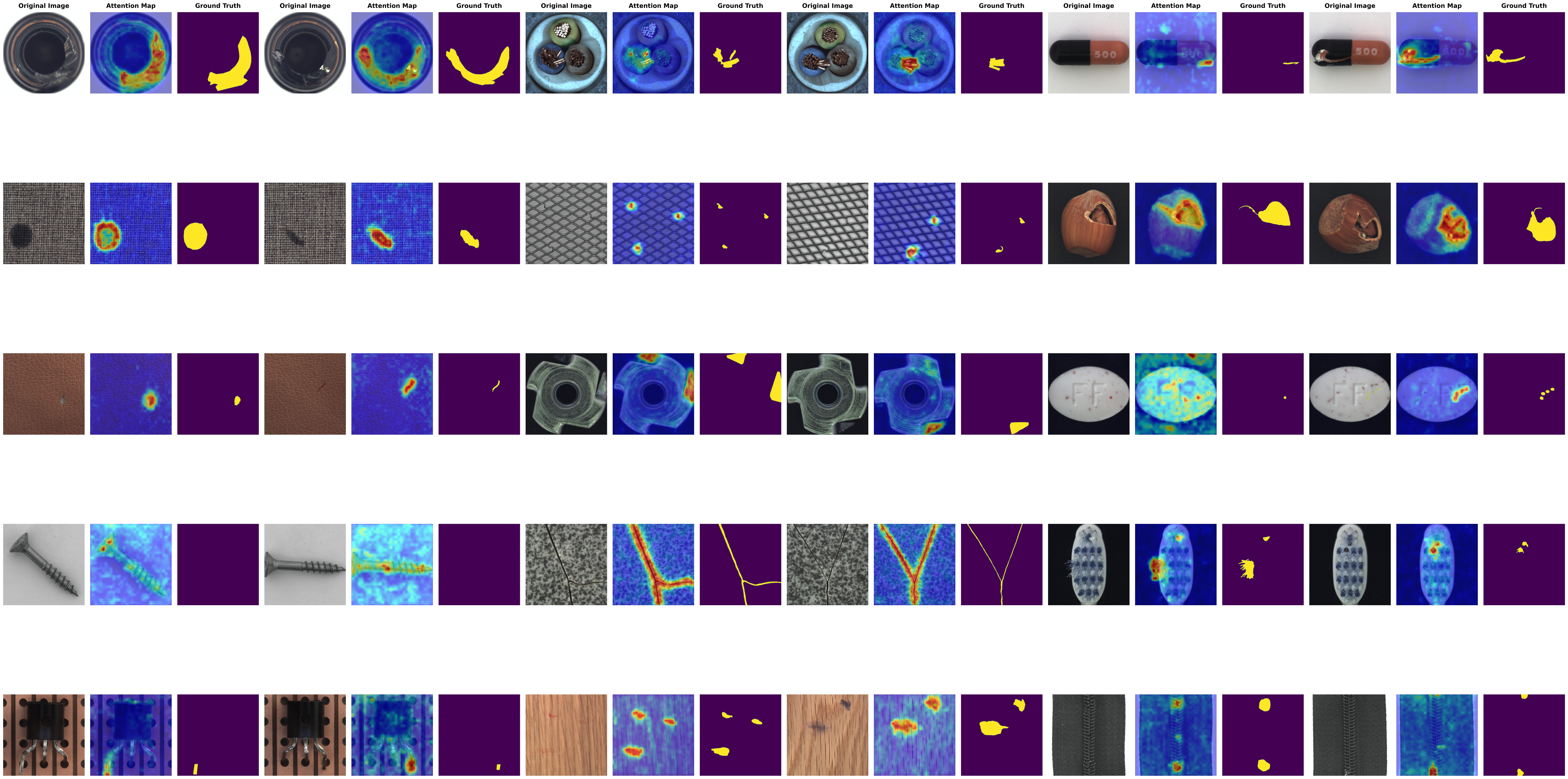}	
	\caption{Qualitative results of our method on MVTec-AD \cite{bergmann2019mvtec}.
	}
	\label{fig_mvtec_visual}
\end{figure*}

\begin{figure*}[t]
	\centering
	\includegraphics[width=0.95 \linewidth]{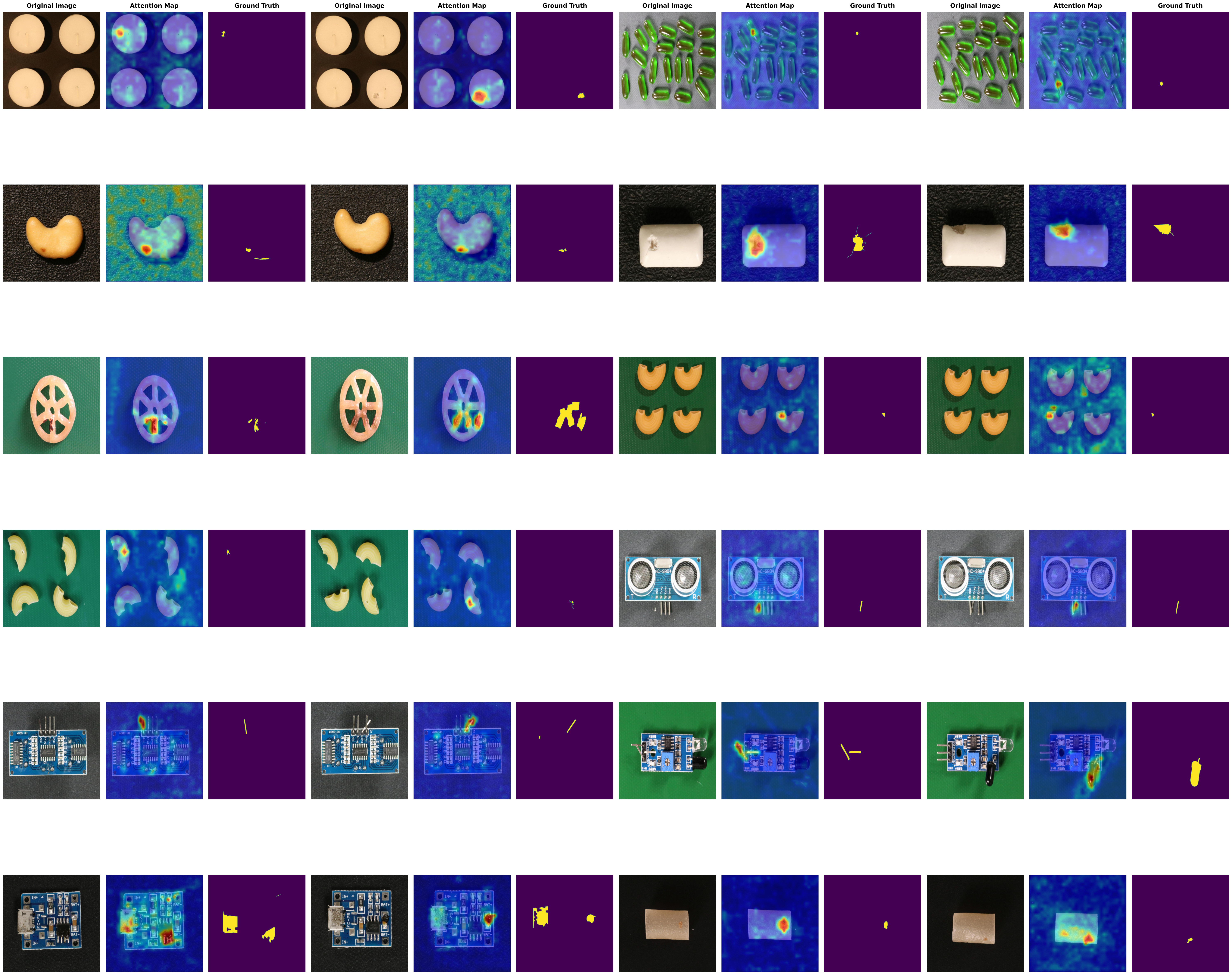}	
	\caption{Qualitative results of our method on ViSA \cite{zou2022spot}.
	}
	\label{fig_mvtec_visa}
\end{figure*}

\begin{figure*}[t]
	\centering
	\includegraphics[width=0.95 \linewidth]{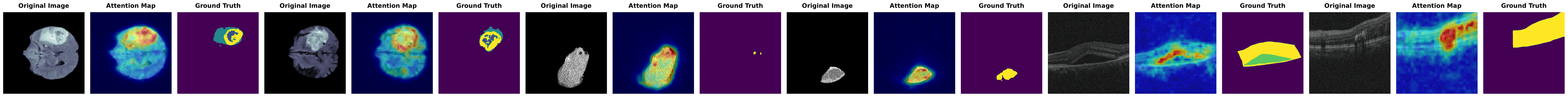}	
	\caption{Qualitative results of our method on Universal Medical \cite{he2024diffusion}.
	}
	\label{fig_mvtec_universal_medical}
\end{figure*}

	We further display the detailed quantitative results on four benchmarks, i.e., MvTec-AD \cite{bergmann2019mvtec}, ViSA \cite{zou2022spot}, Universal Medical \cite{he2024diffusion} and Real-IAD \cite{wang2024real} in Table~\ref{appendix:mvtec}-\ref{appendix:real-iad}, respectively. Also, we display the visualized results of our method on above four datasets in Fig.~\ref{fig_mvtec_visual}-\ref{fig_mvtec_real_iad}, respectively.
	
	\begin{table}[]
		\centering
		\caption{Detailed quantitative results on MvTec-AD \cite{bergmann2019mvtec}. Good score means the averaged anomaly score of normal data in this class, Ungood score indicates that of abnormal data. (\%)}
		\begin{tabular}{l|cccc}
			\toprule
			Class      & I-Auroc/I-AP/I-F1 & P-AUROC/P-AP/P-F1/P-AUPRO & Good Score & UnGood Score \\ \midrule
			carpet     & 99.4/99.8/99.4    & 99.3/71.6/72.6/97.6       & 0.1007     & 0.2806       \\
			grid       & 99.9/100.0/99.1   & 99.5/63.6/62.9/97.7       & 0.0784     & 0.3184       \\
			leather    & 100.0/100.0/100.0 & 99.4/53.2/55.7/98.0       & 0.0795     & 0.2703       \\
			tile       & 100.0/100.0/100.0 & 97.8/73.8/74.7/88.5       & 0.0976     & 0.3215       \\
			wood       & 99.9/100.0/99.2   & 97.7/73.7/70.0/94.2       & 0.1092     & 0.3011       \\
			bottle     & 100.0/100.0/100.0 & 99.2/89.0/84.3/97.0       & 0.0902     & 0.3085       \\
			cable      & 100.0/100.0/99.5  & 98.9/79.9/77.9/94.7       & 0.1169     & 0.2498       \\
			capsule    & 98.4/99.7/98.6    & 98.7/63.5/61.2/97.5       & 0.0825     & 0.1764       \\
			hazelnut   & 100.0/100.0/100.0 & 99.5/83.6/77.7/97.2       & 0.1018     & 0.2503       \\
			metal\_nut & 100.0/100.0/100.0 & 97.0/79.7/86.3/94.8       & 0.1052     & 0.2711       \\
			pill       & 99.7/99.9/98.9    & 97.8/76.1/71.3/97.8       & 0.0831     & 0.1682       \\
			screw      & 99.5/99.8/97.9    & 99.7/65.0/62.1/98.7       & 0.0875     & 0.1494       \\
			toothbrush & 100.0/100.0/100.0 & 99.1/58.9/65.4/95.8       & 0.1115     & 0.2779       \\
			transistor & 99.5/99.2/98.8    & 94.3/60.9/59.2/88.0       & 0.108      & 0.23         \\
			zipper     & 100.0/100.0/100.0 & 99.2/80.5/76.1/97.0       & 0.0731     & 0.2373       \\ \midrule
			Mean       & 99.8/99.9/99.4    & 98.5/71.5/70.5/95.6       & 0.095      & 0.254        \\ \bottomrule
		\end{tabular}
		\label{appendix:mvtec}
	\end{table}
	
	\begin{table}[]
		\centering
		\caption{Detailed quantitative results on VisA \cite{zou2022spot}. Good score means the averaged anomaly score of normal data in this class, Ungood score indicates that of abnormal data. (\%)(\%)}
		\begin{tabular}{l|cccc}
			\toprule
			Class       & I-Auroc/I-AP/I-F1 & P-AUROC/P-AP/P-F1/P-AUPRO & Good Score & UnGood Score \\ \midrule
			candle      & 98.2/98.3/93.5    & 99.4/46.3/49.4/94.9       & 0.0646     & 0.1215       \\
			capsules    & 98.9/99.2/96.9    & 99.6/67.9/67.5/97.8       & 0.0796     & 0.1632       \\
			cashew      & 98.4/99.3/96.5    & 97.5/67.6/63.9/92.2       & 0.0846     & 0.1387       \\
			chewinggum  & 99.7/99.9/98.5    & 99.1/70.5/68.2/87.0       & 0.0894     & 0.2015       \\
			fryum       & 98.7/99.4/96.0    & 97.0/52.5/54.4/93.7       & 0.0774     & 0.164        \\
			macaroni1   & 97.8/97.7/92.5    & 99.5/34.8/40.9/96.0       & 0.0602     & 0.0956       \\
			macaroni2   & 95.6/95.4/89.5    & 99.7/25.3/37.2/98.6       & 0.0662     & 0.0894       \\
			pcb1        & 99.1/99.1/97.1    & 99.7/88.0/79.5/96.1       & 0.0833     & 0.1827       \\
			pcb2        & 99.2/99.0/97.5    & 99.1/29.7/37.3/92.8       & 0.0933     & 0.187        \\
			pcb3        & 99.3/99.3/96.5    & 99.0/42.2/41.3/94.7       & 0.0863     & 0.1801       \\
			pcb4        & 99.9/99.9/98.0    & 98.9/60.4/55.3/94.8       & 0.0827     & 0.2233       \\
			pipe\_fryum & 98.9/99.5/96.4    & 99.2/61.7/65.4/95.3       & 0.0787     & 0.1623       \\ \midrule
			Mean        & 98.7/98.8/95.8    & 99.0/53.9/55.0/94.5       & 0.0789     & 0.1591       \\ \bottomrule
		\end{tabular}
	\end{table}
	\label{appendix:visa}

\begin{table}[]
	\centering
	\caption{Detailed quantitative results on Universal Medical \cite{he2024diffusion}. Good score means the averaged anomaly score of normal data in this class, Ungood score indicates that of abnormal data. (\%)}
	\begin{tabular}{l|cccc}
		\toprule
		Class   & I-Auroc/I-AP/I-F1 & P-AUROC/P-AP/P-F1/P-AUPRO & Good Score & UnGood Score \\ \midrule
		brain   & 95.6/99.0/95.0    & 98.0/69.2/68.3/84.5       & 0.2305     & 0.3586       \\
		liver   & 73.1/68.9/67.0    & 97.5/27.1/33.9/93.5       & 0.198      & 0.2494       \\
		retinal & 94.2/93.0/85.5    & 95.9/71.7/64.7/84.0       & 0.1751     & 0.2837       \\ \midrule
		Mean    & 87.6/87.0/82.5    & 97.1/56.0/55.6/87.3       & 0.2012     & 0.2972       \\ \bottomrule
	\end{tabular}
\end{table}

\begin{table}[]
	\centering
	\caption{Detailed quantitative results on Real-IAD \cite{wang2024real}. Good score means the averaged anomaly score of normal data in this class, Ungood score indicates that of abnormal data. (\%)}
	\begin{tabular}{l|cccc}
		\toprule
		Class               & I-Auroc/I-AP/I-F1 & P-AUROC/P-AP/P-F1/P-AUPRO & Good Score & UnGood Score \\ \midrule
		audiojack           & 90.2/86.1/76.0    & 99.3/51.3/53.8/95.4       & 0.1915     & 0.2997       \\
		bottle\_cap         & 90.5/88.2/80.5    & 99.7/38.9/40.4/98.0       & 0.1167     & 0.2108       \\
		button\_battery     & 89.3/91.3/84.0    & 99.2/60.9/60.5/93.9       & 0.1605     & 0.2796       \\
		end\_cap            & 86.9/86.5/84.2    & 99.0/20.1/31.0/95.8       & 0.165      & 0.217        \\
		eraser              & 93.5/91.6/82.3    & 99.7/47.8/51.2/98.1       & 0.141      & 0.2251       \\
		fire\_hood          & 84.6/76.9/70.6    & 99.4/41.2/45.3/94.8       & 0.167      & 0.234        \\
		mint                & 78.5/78.2/70.4    & 97.8/31.9/40.4/82.2       & 0.1518     & 0.1937       \\
		mounts              & 87.2/76.3/76.2    & 99.5/43.7/45.7/95.2       & 0.1592     & 0.2511       \\
		pcb                 & 93.5/95.9/89.0    & 99.4/60.3/59.9/96.1       & 0.1514     & 0.2496       \\
		phone\_battery      & 93.5/91.6/84.2    & 95.9/43.5/48.8/97.2       & 0.166      & 0.289        \\
		plastic\_nut        & 89.8/83.6/75.8    & 99.7/42.3/47.0/97.8       & 0.1313     & 0.204        \\
		plastic\_plug       & 89.4/86.2/75.8    & 99.2/30.4/37.1/95.4       & 0.1446     & 0.2088       \\
		porcelain\_doll     & 87.1/78.2/71.5    & 99.2/37.5/42.4/95.8       & 0.1317     & 0.1858       \\
		regulator           & 84.2/74.7/64.0    & 99.3/41.5/48.2/95.8       & 0.1294     & 0.1827       \\
		rolled\_strip\_base & 99.0/99.5/96.7    & 99.8/49.7/50.4/98.6       & 0.1235     & 0.2445       \\
		sim\_card\_set      & 96.2/96.4/90.3    & 99.2/57.6/57.0/93.6       & 0.1405     & 0.2615       \\
		switch              & 97.9/98.3/93.3    & 97.3/70.8/66.7/96.3       & 0.1314     & 0.2695       \\
		tape                & 97.0/95.1/89.3    & 99.8/56.0/56.7/98.9       & 0.1154     & 0.2409       \\
		terminalblock       & 98.4/98.9/94.9    & 99.9/54.9/55.5/99.1       & 0.1225     & 0.2376       \\
		toothbrush          & 90.3/92.0/83.3    & 97.1/38.9/44.7/90.4       & 0.1802     & 0.2833       \\
		toy                 & 88.6/91.1/84.6    & 96.5/27.3/34.1/93.3       & 0.1703     & 0.2312       \\
		toy\_brick          & 80.7/76.7/69.2    & 97.6/32.2/39.4/83.8       & 0.2232     & 0.2818       \\
		transistor1         & 97.0/97.9/92.5    & 99.5/54.7/54.2/97.6       & 0.1453     & 0.2863       \\
		usb                 & 95.1/94.6/87.9    & 99.5/49.2/51.6/98.1       & 0.1456     & 0.2786       \\
		usb\_adaptor        & 85.1/78.5/72.4    & 99.2/28.1/37.5/94.2       & 0.1352     & 0.1853       \\
		u\_block            & 92.8/89.1/79.1    & 99.6/49.9/55.6/97.1       & 0.133      & 0.2178       \\
		vcpill              & 93.0/92.7/84.2    & 99.2/73.9/70.0/94.9       & 0.1607     & 0.2727       \\
		wooden\_beads       & 89.0/87.5/79.0    & 99.2/52.2/53.6/92.6       & 0.1695     & 0.2397       \\
		woodstick           & 84.9/74.6/67.5    & 99.2/52.2/54.3/92.3       & 0.2143     & 0.3009       \\
		zipper              & 99.0/99.5/96.4    & 99.0/57.9/59.2/96.5       & 0.139      & 0.3556       \\ \midrule
		Mean                & 90.8/88.2/81.5    & 98.9/46.6/49.7/95.0       & 0.1519     & 0.2473       \\ \bottomrule
	\end{tabular}
	\label{appendix:real-iad}
\end{table}

\begin{figure*}[t]
	\centering
	\includegraphics[width=0.95 \linewidth]{./sec/new_figure/case_visualization_realiad_2.jpg}	
	\caption{Qualitative results of our method on Real-IAD \cite{wang2024real}.
	}
	\label{fig_mvtec_real_iad}
\end{figure*}


\end{document}